COMPUTER SCIENCE

# Challenges of Big Data analysis

Jianqing Fan[1,*], Fang Han[2] and Han Liu[1]

**ABSTRACT**

Big Data bring new opportunities to modern society and challenges to data scientists. On the one hand, Big Data hold great promises for discovering subtle population patterns and heterogeneities that are not possible with small-scale data. On the other hand, the massive sample size and high dimensionality of Big Data introduce unique computational and statistical challenges, including scalability and storage bottleneck, noise accumulation, spurious correlation, incidental endogeneity and measurement errors. These challenges are distinguished and require new computational and statistical paradigm. This paper gives overviews on the salient features of Big Data and how these features impact on paradigm change on statistical and computational methods as well as computing architectures. We also provide various new perspectives on the Big Data analysis and computation. In particular, we emphasize on the viability of the sparsest solution in high-confidence set and point out that exogenous assumptions in most statistical methods for Big Data cannot be validated due to incidental endogeneity. They can lead to wrong statistical inferences and consequently wrong scientific conclusions.

**Keywords:** Big Data, noise accumulation, spurious correlation, incidental endogeneity, data storage, scalability

[1]Department of Operations Research and Financial Engineering, Princeton University, Princeton, NJ 08544, USA and [2]Department of Biostatistics, Johns Hopkins University, Baltimore, MD 21205, USA

*Corresponding author. E-mail: jqfan@princeton.edu

## INTRODUCTION

Big Data promise new levels of scientific discovery and economic value. What is new about Big Data and how they differ from the traditional small- or medium-scale data? This paper overviews the opportunities and challenges brought by Big Data, with emphasis on the distinguished features of Big Data and statistical and computational methods as well as computing architecture to deal with them.

## BACKGROUND

We are entering the era of Big Data—a term that refers to the explosion of available information. Such a Big Data movement is driven by the fact that massive amounts of very high-dimensional or unstructured data are continuously produced and stored with much cheaper cost than they used to be. For example, in genomics we have seen a dramatic drop in price for whole genome sequencing [1]. This is also true in other areas such as social media analysis, biomedical imaging, high-frequency finance, analysis of surveillance videos and retail sales. The existing trend that data can be produced and stored more massively and cheaply is likely to maintain or even accelerate in the future [2]. This trend will have deep impact on science, engineering and business. For example, scientific advances are becoming more and more data-driven and researchers will more and more think of themselves as consumers of data. The massive amounts of high-dimensional data bring both opportunities and new challenges to data analysis. Valid statistical analysis for Big Data is becoming increasingly important.

## GOALS AND CHALLENGES OF ANALYZING BIG DATA

What are the goals of analyzing Big Data? According to [3], two main goals of high-dimensional data analysis are to develop effective methods that can accurately predict the future observations and at the same time to gain insight into the relationship between the features and response for scientific purposes. Furthermore, due to large sample size, Big Data give rise to two additional goals: to understand heterogeneity and commonality across different subpopulations. In other words, Big Data give promises for: (i) exploring the hidden structures of each subpopulation of the data, which is traditionally not feasible and might even be treated as 'outliers' when the sample size is small; (ii) extracting important common features across many subpopulations even when there are large individual variations.

What are the challenges of analyzing Big Data? Big Data are characterized by high dimensionality and large sample size. These two features raise three unique challenges: (i) high dimensionality brings noise accumulation, spurious correlations and incidental homogeneity; (ii) high dimensionality combined with large sample size creates issues such as heavy computational cost and algorithmic instability; (iii) the massive samples in Big Data are typically aggregated from multiple sources at different time points using different technologies. This creates issues of heterogeneity, experimental variations and statistical biases, and requires us to develop more adaptive and robust procedures.

## PARADIGM SHIFTS

To handle the challenges of Big Data, we need new statistical thinking and computational methods. For example, many traditional methods that perform well for moderate sample size do not scale to massive data. Similarly, many statistical methods that perform well for low-dimensional data are facing significant challenges in analyzing high-dimensional data. To design effective statistical procedures for exploring and predicting Big Data, we need to address Big Data problems such as heterogeneity, noise accumulation, spurious correlations and incidental endorgeneity, in addition to balancing the statistical accuracy and computational efficiency.

In terms of statistical accuracy, dimension reduction and variable selection play pivotal roles in analyzing high-dimensional data. This is designed to address noise accumulation issues. For example, in high-dimensional classification, [4] and [5] showed that conventional classification rules using all features perform no better than random guess due to noise accumulation. This motivates new regularization methods [6–10] and sure independence screening [11–13]. Furthermore, high dimensionality introduces spurious correlations between response and unrelated covariates, which may lead to wrong statistical inference and false scientific conclusions [14]. High dimensionality also gives rise to incidental endogeneity, a phenomenon that many unrelated covariates may incidentally be correlated with the residual noises. The endogeneity creates statistical biases and causes model selection inconsistency that lead to wrong scientific discoveries [15,16]. Yet, most statistical procedures are based on unrealistic exogenous assumptions that cannot be validated by data (see the 'Incidental endogeneity' section and [17]). New statistical procedures with these issues in mind are crucially needed.

In terms of computational efficiency, Big Data motivate the development of new computational infrastructure and data-storage methods. Optimization is often a tool, not a goal, to Big Data analysis. Such a paradigm change has led to significant progresses on developments of fast algorithms that are scalable to massive data with high dimensionality. This forges cross-fertilizations among different fields including statistics, optimization and applied mathematics. For example, the authors of [18] showed that the non-deterministic polynomial-time hard (NP-hard) best subset regression can be recast as an $L_1$-norm penalized least-squares problem which can be solved by the interior point method. Alternative algorithms to accelerate this $L_1$-norm penalized least-squares problems, such as least angle regression [19], threshold gradient descent [20] and coordinate descent [21,22], iterative shrinkage-thresholding algorithms [23,24], are proposed. Besides large-scale optimization algorithms, Big Data also motivate the development of majorization–minimization algorithms [25–27], 'large-scale screening and small-scale optimization' framework [28], parallel computing methods [29–31] and approximate algorithms that are scalable to large sample size.

## ORGANIZATION OF THIS PAPER

The rest of this paper is organized as follows. The section 'Rises of Big Data' overviews the rise of Big Data problem from science, engineering and social science. The 'Salient Features of Big Data' section explains some unique features of Big Data and their impacts on statistical inference. Statistical methods that tackle these Big Data problems are given in the 'Impact on statistical thinking' section. The 'Impact on computing infrastructure' section gives an overview on scalable computing infrastructure for Big Data storage and processing. The 'Impact on computational methods' section discusses the computational aspect of Big Data and introduces some recent progresses. The 'Conclusions and future perspectives' section concludes the paper.

## RISE OF BIG DATA

Massive sample size and high dimensionality characterize many contemporary datasets. For example,

in genomics, there have been more than 500 000 microarrays that are publicly available with each array containing tens of thousands of expression values of molecules; in biomedical engineering, there have been tens of thousands of terabytes of functional magnetic resonance images (fMRIs) with each image containing more than 50 000 voxel values. Other examples of massive and high-dimensional data include unstructured text corpus, social medias, and financial time series, e-commerce data, retail transaction records and surveillance videos. We now briefly illustrate some of these Big Data problems.

### Genomics

Many new technologies have been developed in genomics and enable inexpensive and high-throughput measurement of the whole genome and transcriptome. These technologies allow biologists to generate hundreds of thousands of datasets and have shifted their primary interests from the acquisition of biological sequences to the study of biological function. The availability of massive datasets sheds light towards new scientific discoveries. For example, the large amount of genome sequencing data now make it possible to uncover the genetic markers of rare disorders [32,33] and find associations between diseases and rare sequence variants [34,35]. The breakthroughs in biomedical imaging technology allow scientists to simultaneously monitor many gene and protein functions, permitting us to study interactions in regulatory processes and neuron activities. Moreover, the emergence of publicly available genomic databases enables integrative analysis which combines information from many sources for drawing scientific conclusions. These research studies give rise to many computational methods as well as new statistical thinking and challenges [36].

One of the important steps in genomic data analysis is to remove systematic biases (e.g. intensity effect, batch effect, dye effect, block effect, among others). Such systematic biases are due to experimental variations, such as environmental, demographic, and other technical factors, and can be more severe when we combine different data sources. They have been shown to have substantial effects on gene expression levels, and failing to taking them into consideration may lead to wrong scientific conclusions [37]. When the data are aggregated from multiple sources, it remains an open problem on what is the best normalization practice.

Even with the systematic biases removed, another challenge is to conduct large-scale tests to pick important genes, proteins, or single-nucleotide polymorphism (SNP). In testing the significance of thousands of genes, classical methods of controlling the probability of making one falsely discovered gene are no longer suitable and alternative procedures have been designed to control the false discovery rates [38–42] and to improve the power of the tests [42]. These technologies, though high-throughput in measuring the expression levels of tens of thousands of genes, remain low-throughput in surveying biological contexts (e.g. novel cell types, tissues, diseases, etc.).

An additional challenge in genomic data analysis is to model and explore the underlying heterogeneity of the aggregated datasets. Due to technology limitations and resource constraints, a single lab usually can only afford performing experiments for no more than a few cell types. This creates a major barrier for comprehensively characterizing gene regulation in all biological contexts, which is a fundamental goal of functional genomics. On the other hand, the National Center for Biotechnology Information (NCBI) Gene Expression Omnibus (GEO) [43] and other public databases have cumulated more than 500 000 gene expression profiles, including microarray, exon array and ribonucleic acid-sequencing (RNA-seq) samples from thousands of biological contexts. Public ChIP–chip and ChIP–seq data generated by different labs for different proteins and in different contexts are also steadily growing. Together, these public data contain enormous amounts of information that have not been fully exploited so far. Massive data aggregated from these public databases shed light on systematically studying many biological contexts in a high-throughput way. However, how to systematically explore the underlying heterogeneity and unveil the commonality across different subpopulations remains an active research area.

### Neuroscience

Many important diseases, including Alzheimer's disease, Schizophrenia, Attention Deficit Hyperactive Disorder, Depression and Anxiety, have been shown to be related to brain connectivity networks. Understanding the hierarchical, complex, functional network organization of the brain is a necessary first step to explore how the brain changes with disease. Rapid advances in neuroimaging techniques, such as fMRI and positron emission tomography as well as electrophysiology, provide great potential for the study of functional brain networks, i.e. the coherence of the activities among different brain regions [44].

Take fMRI for example. It is a non-invasive technique for determining the neural correlates of mental processes in humans. During the past decade, this technique has become a leading method in the fields of cognitive and physiological neuroscience and kept producing massive amounts of high-resolution brain images. These images enable us to explore the association between brain connectivity and potential responses such as disease or psychological status. The fMRI data are massive and very high dimensional. Due to its non-invasive feature, everyday many fMRI machines keep scanning different subjects and constantly produce new imaging data. For each data point, the subject's brain is scanned for hundreds of times. Therefore, it is a 3D time-course image which contains more than hundreds of thousands of voxels. At the same time, the fMRI images are noisy due to its technological limit and possible head motion of the subjects. Analyzing such high-dimensional and noisy data poses great challenges to statisticians and neuroscientists.

Similar to the field of genomics, an important Big Data problem in neuroscience is to aggregate datasets from multiple sources. Brain imaging data sharing is becoming more and more frequent nowadays [45]. Primary sources of fMRI data arise from the International Data Sharing Initiative and the 1000 Functional Connectomes Project [46], Autism Brain Imaging Data Exchange (ABIDE) [47] and ADHD-200 [48] datasets. These international efforts have compiled thousands of resting-state fMRI scans along with complimentary structural scans. The largest of the datasets is the 1000 Functional Connectomes Project, which focuses on healthy adults and includes limited covariate information on age, gender, handedness and image quality. The ADHD-200 dataset is similarly structured; yet, it includes diagnostic information on disease status such as human IQ. The ABIDE dataset is similar to the ADHD-200 dataset, with diagnostic autism and symptom severity information. However, it has a greater balance between diseased and non-diseased subjects. These large datasets pose great opportunities as well as new challenges.

One of the main challenges, as in the area of genomics, is to remove the systematic biases caused by experimental variations and data aggregations. Moreover, statistically controlled inclusion of a subject in a group study, i.e. testing whether a person should be rejected as outlier data, is often poorly conducted [49] and voxels cannot be perfectly aligned across different experiments in different laboratories. Therefore, the collected data contain many outliers and missing values. These issues make data preprocessing and analysis significantly more complicated. Many traditional statistical procedures are not well suited in this noisy high-dimensional settings, and new statistical thinking is crucially needed.

## Economics and finance

Over the past decade, more and more corporations are adopting the data-driven approach to conduct more targeted services, reduce risks and improve performance. They are implementing specialized data analytics programs to collect, store, manage and analyze large datasets from a range of sources to identify key business insights that can be exploited to support better decision making. For example, available financial data sources include stock prices, currency and derivative trades, transaction records, high-frequency trades, unstructured news and texts, consumers' confidence and business sentiments buried in social media and internet, among others. Analyzing these massive datasets helps measuring firms risks as well as systematic risks. It requires professionals who are familiar with sophisticated statistical techniques in portfolio management, securities regulation, proprietary trading, financial consulting and risk management.

Analyzing a large panel of economic and financial data is challenging. For example, as an important tool in analyzing the joint evolution of macroeconomics time series, the conventional vector autoregressive (VAR) model usually includes no more than 10 variables, given the fact that the number of parameters grows quadratically with the size of the model. However, nowadays econometricians need to analyze multivariate time series with more than hundreds of variables. Incorporating all information into the VAR model will cause severe overfitting and bad prediction performance. One solution is to resort to sparsity assumptions, under which new statistical tools have been developed [50,51].

Another example is portfolio optimization and risk management [52,53]. In this problem, estimating the covariance and inverse covariance matrices of the returns of the assets in the portfolio plays an important role. Suppose that we have 1000 stocks to be managed. There are 500 500 covariance parameters to be estimated. Even if we could estimate each individual parameter accurately, the cumulated error of the whole matrix estimation can be large under matrix norms. This requires new statistical procedures. See, for example, [54–66] on estimating large covariance matrices and their inverse.

## Other applications

Big Data have numerous other applications. Taking social network data analysis for example, massive amount of social network data are being produced by Twitter, Facebook, LinkedIn and YouTube. These data reveal numerous individual's characteristics and have been exploited in various fields. For example, the authors of [67] used these data to predict influenza epidemic; those of [68] used these data to predict the stock market trend; and the authors of [69] used the social network data to predict box-office revenues for movies. In addition, the social media and Internet contain massive amount of information on the consumer preferences and confidences, leading economics indicators, business cycles, political attitudes, and the economic and social states of a society. It is anticipated that the social network data will continue to explode and be exploited for many new applications.

Several other new applications that are becoming possible in the Big Data era include:

(i) *Personalized services.* With more personal data collected, commercial enterprises are able to provide personalized services adapt to individual preferences. For example, Target (a retailing company in the United States) is able to predict a customer's need by analyzing the collected transaction records.

(ii) *Internet security.* When a network-based attack takes place, historical data on network traffic may allow us to efficiently identify the source and targets of the attack.

(iii) *Personalized medicine.* More and more health-related metrics such as individual's molecular characteristics, human activities, human habits and environmental factors are now available. Using these pieces of information, it is possible to diagnose an individual's disease and select individualized treatments.

(iv) *Digital humanities.* Nowadays many archives are being digitized. For example, Google has scanned millions of books and identified about every word in every one of those books. This produces massive amount of data and enables addressing topics in the humanities, such as mapping the transportation system in ancient Roman, visualizing the economic connections of ancient China, studying how natural languages evolve over time, or analyzing historical events.

## SALIENT FEATURES OF BIG DATA

Big Data create unique features that are not shared by the traditional datasets. These features pose significant challenges to data analysis and motivate the development of new statistical methods. Unlike traditional datasets where the sample size is typically larger than the dimension, Big Data are characterized by massive sample size and high dimensionality. First, we will discuss the impact of large sample size on understanding heterogeneity: on the one hand, massive sample size allows us to unveil hidden patterns associated with small subpopulations and weak commonality across the whole population. On the other hand, modeling the intrinsic heterogeneity of Big Data requires more sophisticated statistical methods. Secondly, we discuss several unique phenomena associated with high dimensionality, including noise accumulation, spurious correlation and incidental endogeneity. These unique features make traditional statistical procedures inappropriate. Unfortunately, most high-dimensional statistical techniques address only noise accumulation and spurious correlations issues, but not incidental endogeneity. They are based on exogeneity assumptions that often cannot be validated by collected data, due to incidental endogeneity.

## Heterogeneity

Big Data are often created via aggregating many data sources corresponding to different subpopulations. Each subpopulation might exhibit some unique features not shared by others. In classical settings where the sample size is small or moderate, data points from small subpopulations are generally categorized as 'outliers', and it is hard to systematically model them due to insufficient observations. However, in the Big Data era, the large sample size enables us to better understand heterogeneity, shedding light toward studies such as exploring the association between certain covariates (e.g. genes or SNPs) and rare outcomes (e.g. rare diseases or diseases in small populations) and understanding why certain treatments (e.g. chemotherapy) benefit a subpopulation and harm another subpopulation. To better illustrate this point, we introduce the following mixture model for the population:

$$\lambda_1 p_1(y; \boldsymbol{\theta}_1(\mathbf{x})) + \cdots + \lambda_m p_m(y; \boldsymbol{\theta}_m(\mathbf{x})), \quad (1)$$

where $\lambda_j \geq 0$ represents the proportion of the $j$th subpopulation, $p_j(y; \boldsymbol{\theta}_j(\mathbf{x}))$ is the probability distribution of the response of the $j$th subpopulation given the covariates $\mathbf{x}$ with $\boldsymbol{\theta}_j(\mathbf{x})$ as the parameter vector. In practice, many subpopulations are rarely observed, i.e. $\lambda_j$ is very small. When the sample size $n$ is moderate, $n\lambda_j$ can be small, making it infeasible to infer the covariate-dependent parameters $\boldsymbol{\theta}_j(\mathbf{x})$ due to the lack of information. However, because Big Data are characterized by large sample size $n$, the sample size $n\lambda_j$ for the $j$th subpopulation can be moderately large even if $\lambda_j$ is very small. This enables us to more accurately infer about the subpopulation parameters $\boldsymbol{\theta}_j(\cdot)$. In short, the main advantage brought by Big Data is to understand the heterogeneity of subpopulations, such as the benefits of certain personalized treatments, which are infeasible when sample size is small or moderate.

Big Data also allow us to unveil weak commonality across whole population, thanks to large sample sizes. For example, the benefit of one drink of red wine per night on heart can be difficult to assess without large sample size. Similarly, health risks to exposure of certain environmental factors can only be more convincingly evaluated when the sample sizes are sufficiently large.

Besides the aforementioned advantages, the heterogeneity of Big Data also poses significant challenges to statistical inference. Inferring the mixture model in (1) for large datasets requires sophisticated statistical and computational methods. In low dimensions, standard techniques such as the expectation–maximization algorithm for finite mixture models can be applied. In high dimensions, however, we need to carefully regularize the estimating procedure to avoid overfitting or noise accumulation and to devise good computation algorithms [70,71].

## Noise accumulation

Analyzing Big Data requires us to simultaneously estimate or test many parameters. These estimation errors accumulate when a decision or prediction rule depends on a large number of such parameters. Such a noise accumulation effect is especially severe in high dimensions and may even dominate the true signals. It is usually handled by the sparsity assumption [2,72,73].

Take high-dimensional classification for instance. Poor classification is due to the existence of many weak features that do not contribute to the reduction of classification error [4]. As an example, we consider a classification problem where the data come from two classes:

$$X_1, \ldots, X_n \sim N_d(\mu_1, I_d)$$
$$\text{and } Y_1, \ldots, Y_n \sim N_d(\mu_2, I_d). \quad (2)$$

We want to construct a classification rule which classifies a new observation $Z \in \mathbb{R}^d$ into either the first or the second class. To illustrate the impact of noise accumulation in classification, we set $n = 100$ and $d = 1000$. We set $\mu_1 = 0$ and $\mu_2$ to be sparse, i.e. only the first 10 entries of $\mu_2$ are nonzero with value 3, and all the other entries are zero. Figure 1 plots the first two principal components by using the first $m = 2, 40, 200$ features and the whole 1000 features. As illustrated in these plots, when $m = 2$ we obtain high discriminative power. However, the discriminative power becomes very low when $m$ is too large due to noise accumulation. The first 10 features contribute to classifications and the remaining features do not. Therefore, when $m > 10$, procedures do not obtain any additional signals, but accumulate noises: the larger the $m$, the more the noise accumulates, which deteriorates the classification procedure with dimensionality. For $m = 40$, the accumulated signals compensate the accumulated noise, so that the first two principal components still have good discriminative power. When $m = 200$, the accumulated noise exceeds the signal gains.

The above discussion motivates the usage of sparse models and variable selection to overcome the effect of noise accumulation. For example, in the classification model (2), instead of using all the features, we could select a subset of features which attain the best signal-to-noise ratio. Such a sparse model provides more improved classification performance [72,73]. In other words, variable selection plays a pivotal role in overcoming noise accumulation in classification and regression prediction. However, variable selection in high dimensions is challenging due to spurious correlation, incidental endogeneity, heterogeneity and measurement errors.

## Spurious correlation

High dimensionality also brings spurious correlation, referring to the fact that many uncorrelated random variables may have high sample correlations in high dimensions. Spurious correlation may cause false scientific discoveries and wrong statistical inferences.

Consider the problem of estimating the coefficient vector $\boldsymbol{\beta}$ of a linear model

$$\mathbf{y} = \mathbf{X}\boldsymbol{\beta} + \boldsymbol{\epsilon}, \quad \text{Var}(\boldsymbol{\epsilon}) = \sigma^2 \mathbf{I}_d, \quad (3)$$

where $\mathbf{y} \in \mathbb{R}^n$ represents the response vector, $\mathbf{X} = [\mathbf{x}_1, \ldots, \mathbf{x}_n]^T \in \mathbb{R}^{n \times d}$ represents the design matrix, $\boldsymbol{\epsilon} \in \mathbb{R}^n$ represents an independent random noise vector and $\mathbf{I}_d$ is the $d \times d$ identity matrix. To cope with the noise accumulation issue, when the dimension $d$ is comparable to or larger than the sample size $n$, it is popular to assume that only a small number of variables contribute to the response, i.e. $\boldsymbol{\beta}$ is a sparse vector. Under this sparsity assumption, variable selection can be conducted to avoid noise accumulation, improve the performance of prediction, as well as enhance the interpretability of the model with parsimonious representation.

In high dimensions, even for a model as simple as (3), variable selection is challenging due to the presence of spurious correlation. In particular, [11] showed that, when the dimensionality is high, the important variables can be highly correlated with several spurious variables which are scientifically unrelated. We consider a simple example to illustrate this phenomenon. Let $\mathbf{x}_1, \ldots, \mathbf{x}_n$ be $n$ independent observations of a $d$-dimensional Gaussian random vector $X = (X_1, \ldots, X_d)^T \sim N_d(\mathbf{0}, \mathbf{I}_d)$. We repeatedly simulate the data with $n = 60$ and $d = 800$ and 6400 for 1000 times. Figure 2a shows the empirical distribution of the maximum absolute sample correlation coefficient between the first variable with the remaining ones defined as

$$\widehat{r} = \max_{j \geq 2} |\widehat{\mathrm{Corr}}(X_1, X_j)|, \qquad (4)$$

 

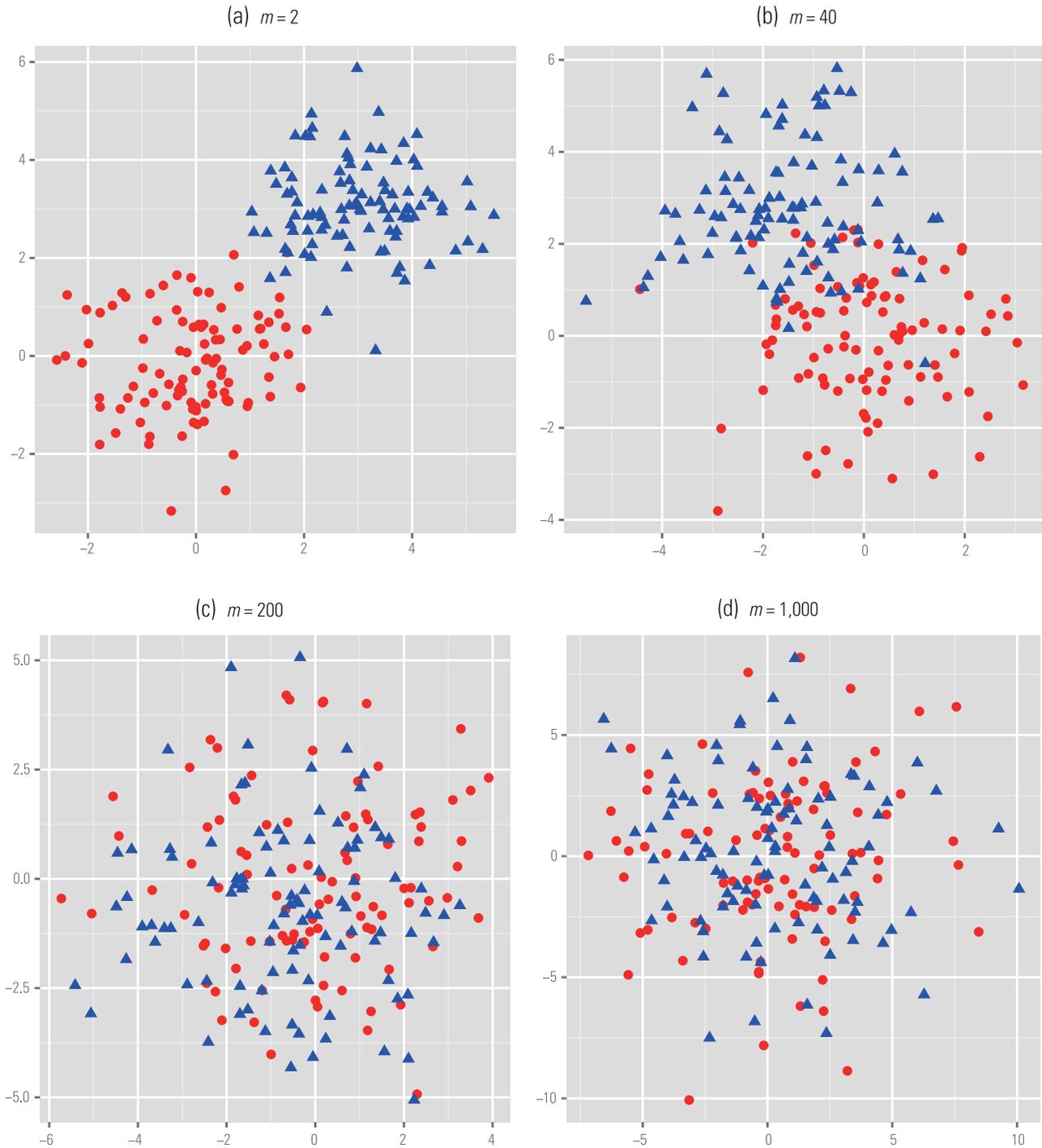

**Figure 1.** Scatter plots of projections of the observed data ($n = 100$ from each class) onto the first two principal components of the best $m$-dimensional selected feature space. A projected data with the filled circle indicates the first class and the filled triangle indicates the second class.

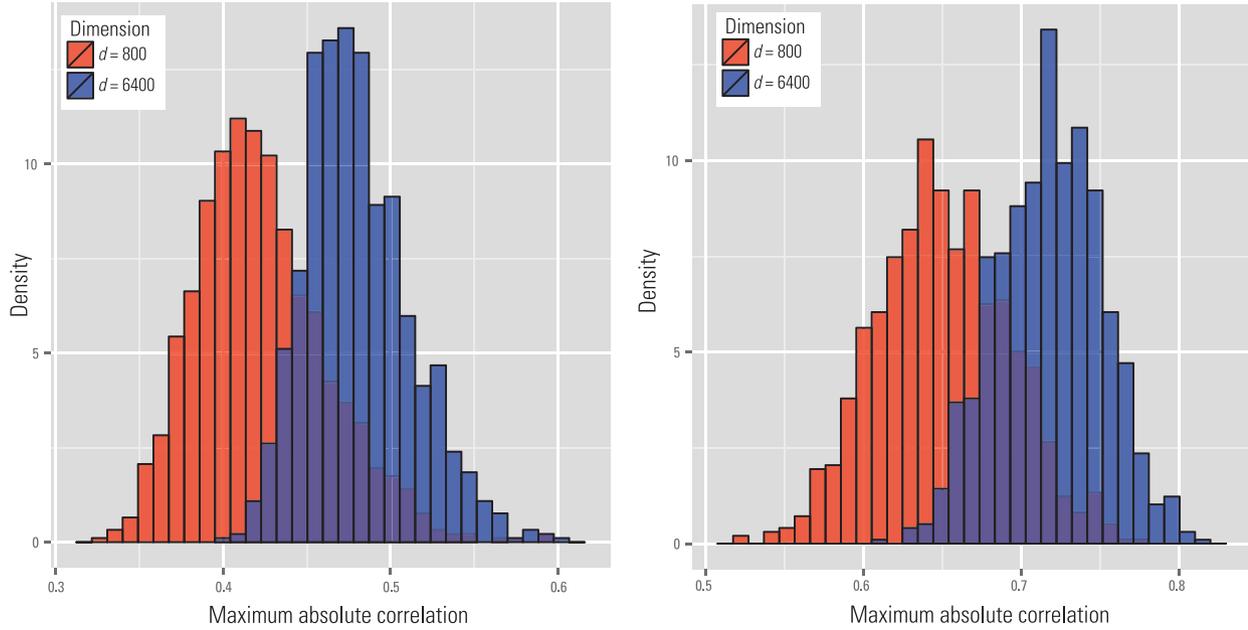

**Figure 2.** Illustration of spurious correlation. (a) Distribution of the maximum absolute sample correlation coefficients between $X_1$ and $\{X_j\}_{j\neq 1}$. (b) Distribution of the maximum absolute sample correlation coefficients between $X_1$ and the closest linear projections of any four members of $\{X_j\}_{j\neq 1}$ to $X_1$. Here the dimension $d$ is 800 and 6400, the sample size $n$ is 60. The result is based on 1000 simulations.

where $\widehat{\mathrm{Corr}}(X_1, X_j)$ is the sample correlation between the variables $X_1$ and $X_j$. We see that the maximum absolute sample correlation becomes higher as dimensionality increases.

Furthermore, we can compute the maximum absolute multiple correlation between $X_1$ and linear combinations of several irrelevant spurious variables:

$$\widehat{R} = \max_{|S|=4} \max_{\{\beta_j\}_{j=1}^4} \left| \widehat{\mathrm{Corr}}\left(X_1, \sum_{j\in S}\beta_j X_j\right)\right|. \quad (5)$$

Using the same configuration as in Fig. 2 a, Fig. 2 b plots the empirical distribution of the maximum absolute sample correlation coefficient between $X_1$ and $\sum_{j\in S}\beta_j X_j$, where $S$ is any size four subset of $\{2, \ldots, d\}$ and $\beta_j$ is the least-squares regression coefficient of $X_j$ when regressing $X_1$ on $\{X_j\}_{j\in S}$. Again, we see that even though $X_1$ is utterly independent of $X_2, \ldots, X_d$, the correlation between $X_1$ and the closest linear combination of any four variables of $\{X_j\}_{j\neq 1}$ to $X_1$ can be very high. We refer to [14] and [74] about more theoretical results on characterizing the orders of $\widehat{r}$.

The spurious correlation has significant impact on variable selection and may lead to false scientific discoveries. Let $X_S = (X_j)_{j\in S}$ be the sub-random vector indexed by $S$ and let $\widehat{S}$ be the selected set that has the higher spurious correlation with $X_1$ as in Fig. 2. For example, when $n=60$ and $d=6400$, we see that $X_1$ is practically indistinguishable from $X_{\widehat{S}}$ for a set $\widehat{S}$ with $|\widehat{S}|=4$. If $X_1$ represents the expression level of a gene that is responsible for a disease, we cannot distinguish it from the other four genes in $\widehat{S}$ that have a similar predictive power although they are scientifically irrelevant.

Besides variable selection, spurious correlation may also lead to wrong statistical inference. We explain this by considering again the same linear model as in (3). Here we would like to estimate the standard error $\sigma$ of the residual, which is prominently featured in statistical inferences of regression coefficients, model selection, goodness-of-fit test and marginal regression. Let $\widehat{S}$ be a set of selected variables and $\mathbf{P}_{\widehat{S}}$ be the projection matrix on the column space of $\mathbf{X}_{\widehat{S}}$. The standard residual variance estimator, based on the selected variables, is

$$\widehat{\sigma}^2 = \frac{\mathbf{y}^T(\mathbf{I}_n - \mathbf{P}_{\widehat{S}})\mathbf{y}}{n - |\widehat{S}|}. \quad (6)$$

The estimator (6) is unbiased when the variables are not selected by data and the model is correct. However, the situation is completely different when the variables are selected based on data. In particular, the authors of [14] showed that when there are many spurious variables, $\sigma^2$ is seriously underestimated, which leads further to wrong statistical inferences including model selection or significance tests, and false scientific discoveries such as finding wrong genes for molecular mechanisms. They also propose a refitted cross-validation method to attenuate the problem.

**Incidental endogeneity**

Incidental endogeneity is another subtle issue raised by high dimensionality. In a regression setting $Y = \sum_{j=1}^{d} \beta_j X_j + \varepsilon$, the term 'endogeneity' [75] means that some predictors $\{X_j\}$ correlate with the residual noise $\varepsilon$. The conventional sparse model assumes

$$Y = \sum_j \beta_j X_j + \varepsilon,$$
$$\text{and } \mathbb{E}(\varepsilon X_j) = 0 \quad \text{for } j = 1, \ldots, d, \quad (7)$$

with a small set $S = \{j: \beta_j \neq 0\}$. The exogenous assumption in (7) that the residual noise $\varepsilon$ is uncorrelated with all the predictors is crucial for validity of most existing statistical procedures, including variable selection consistency. Though this assumption looks innocent, it is easy to be violated in high dimensions as some of variables $\{X_j\}$ are incidentally correlated with $\varepsilon$, making most high-dimensional procedures statistically invalid.

To explain the endogeneity problem in more detail, suppose that unknown to us, the response $Y$ is related to three covariates as follows:

$$Y = X_1 + X_2 + X_3 + \varepsilon,$$
$$\text{with} \quad \mathbb{E}\varepsilon X_j = 0, \quad \text{for } j = 1, 2, 3.$$

In the data-collection stage, we do not know the true model, and therefore collect as many covariates that are potentially related to $Y$ as possible, in hope to include all members in $S$ in (7). Incidentally, some of those $X_j$s (for $j \neq 1, 2, 3$) might be correlated with the residual noise $\varepsilon$. This invalidates the exogenous modeling assumption in (7). In fact, the more covariates are collected or measured, the harder this assumption is satisfied.

Unlike spurious correlation, incidental endogeneity refers to the genuine existence of correlations between variables unintentionally, both due to high dimensionality. The former is analogous to find two persons look alike but have no genetic relation, whereas the latter is similar to bumping into an acquaintance, both easily occurring in a big city. More generally, endogeneity occurs as a result of selection biases, measurement errors and omitted variables. These phenomena arise frequently in the analysis of Big Data, mainly due to two reasons:

- With the benefit of new high-throughput measurement techniques, scientists are able to and tend to collect as many features as possible. This accordingly increases the possibility that some of them might be correlated with the residual noise, incidentally.
- Big Data are usually aggregated from multiple sources with potentially different data generating schemes. This increases the possibility of selection bias and measurement errors, which also cause potential incidental endogeneity.

Whether incidental endogeneity appears in real datasets and how shall we test it in practice? We consider a genomics study in which 148 microarray samples are downloaded from GEO database and ArrayExpress [76]. These samples are created under the Affymetrix HGU133a platform for human subjects with prostate cancer. The obtained dataset contains 22 283 probes, corresponding to 12 719 genes. In this example, we are interested in the gene named 'Discoidin domain receptor family, member 1' (abbreviated as DDR1). DDR1 encodes receptor tyrosine kinases, which plays an important role in the communication of cells with their microenvironment. DDR1 is known to be highly related to the prostate cancer [77] and we wish to study its association with other genes in patients with prostate cancer. We took the gene expressions of DDR1 as the response variable $Y$ and the expressions of all the remaining 12 718 genes as predictors. The left panel of Fig. 3 draws the empirical distribution of the correlations between the response and individual predictors.

To illustrate the existence of endogeneity, we fit an $L_1$-penalized least-squares regression (Lasso) on the data, and the penalty is automatically selected via 10-fold cross validation (37 genes are selected). We then refit an ordinary least-squares regression on the selected model to calculate the residual vector. In the right panel of Fig. 3, we plot the empirical distribution of the correlations between the predictors and the residuals. We see the residual noise is highly correlated with many predictors. To make sure these correlations are not purely caused by spurious correlation, we introduce a 'null distribution' of the spurious correlations by randomly permuting the orders of rows in the design matrix, such that the predictors are indeed independent of the residual noise. By comparing the two distributions, we see that the distribution of correlations between predictors and residual noise on the raw data (labeled 'raw data') has a heavier tail than that on the permuted data (labeled 'permuted data'). This result provides stark evidence of endogeneity in these data.

The above discussion shows that incidental endogeneity is likely to be present in Big Data. The problem of dealing with endogenous variables is not well understood in high-dimensional statistics. What is the consequence of this endogeneity? The authors of [16] showed that endogeneity causes inconsistency in model selection. In particular, they provided thorough analysis to illustrate the impact of endogeneity on high-dimensional statistical inference and proposed alternative methods to conduct linear regression with consistency guarantees under weaker conditions. See also the following section.

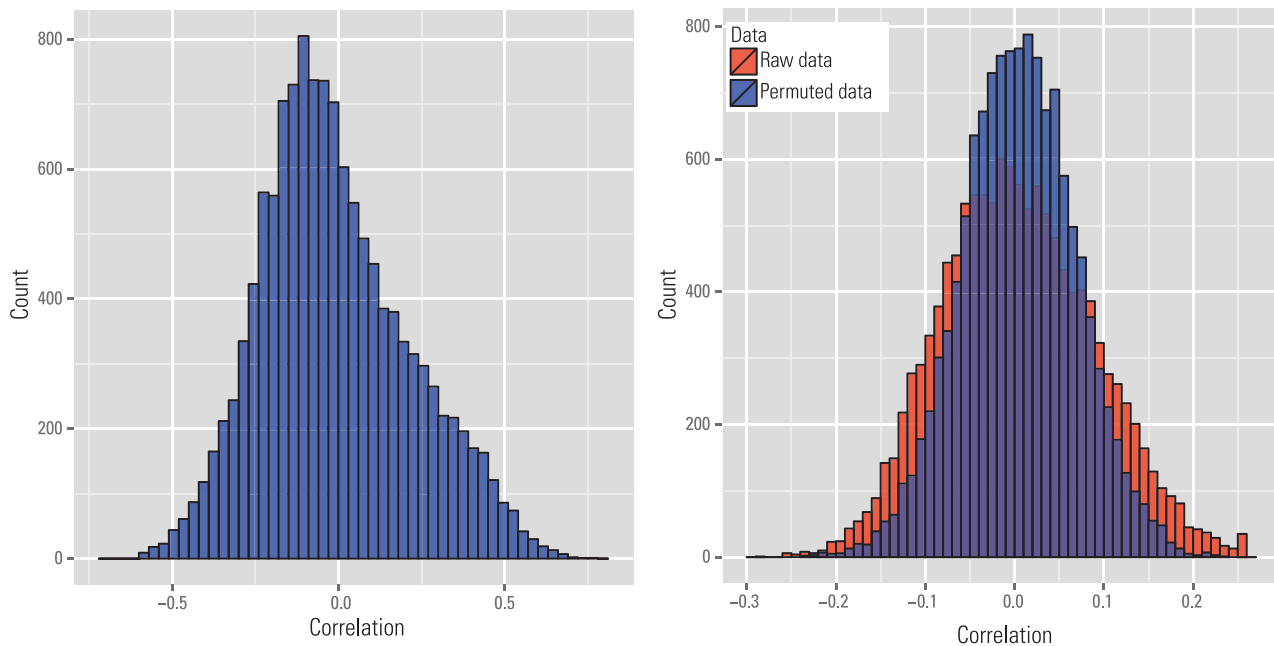

**Figure 3.** Illustration of incidental endogeneity on a microarry gene expression data. Left panel: the distribution of the sample correlation $\widehat{\text{Corr}}(X_j, Y)$ ($j = 1, \ldots, 12\,718$). Right panel: the distribution of the sample correlation $\widehat{\text{Corr}}(X_j, \hat{\varepsilon})$. Here $\hat{\varepsilon}$ represents the residual noise after the Lasso fit. We provide the distributions of the sample correlations using both the raw data and permuted data.

## IMPACT ON STATISTICAL THINKING

As has been shown in the previous section, massive sample size and high dimensionality bring heterogeneity, noise accumulation, spurious correlation and incidental endogeneity. These features of Big Data make traditional statistical methods invalid. In this section, we introduce new statistical methods that can handle these challenges. For an overview, see [72] and [73].

### Penalized quasi-likelihood

To handle the noise-accumulation issue, we assume that the model parameter $\boldsymbol{\beta}$ as in (3) is sparse. The classical model selection theory, according to [78], suggests to choose a parameter vector $\boldsymbol{\beta}$ that minimizes negative penalized quasi-likelihood:

$$-\text{QL}(\boldsymbol{\beta}) + \lambda \|\boldsymbol{\beta}\|_0, \qquad (8)$$

where $\text{QL}(\boldsymbol{\beta})$ is the quasi-likelihood of $\boldsymbol{\beta}$ and $\|\cdot\|_0$ represents the $L_0$ pseudo-norm (i.e. the number of nonzero entries in a vector). Here $\lambda > 0$ is a regularization parameter that controls the bias-variance tradeoff. The solution to the optimization problem in (8) has nice statistical properties [79]. However, it is essentially combinatoric optimization and does not scale to large-scale problems.

The estimator in (8) can be extended to a more general form

$$\ell_n(\boldsymbol{\beta}) + \sum_{j=1}^{d} P_{\lambda,\gamma}(\beta_j), \qquad (9)$$

where the term $\ell_n(\boldsymbol{\beta})$ measures the goodness of fit of the model with parameter $\boldsymbol{\beta}$ and $\sum_{j=1}^{d} P_{\lambda,\gamma}(\beta_j)$ is a sparsity-inducing penalty that encourages sparsity, in which $\lambda$ is again the tuning parameter that controls the bias-variance tradeoff and $\gamma$ is a possible fine-tune parameter which controls the degree of concavity of the penalty function [8]. Popular choices of the penalty function $P_{\lambda,\gamma}(\cdot)$ include the hard-thresholding penalty [80,81], soft-thresholding penalty [6,82], smoothly clipped absolution deviation (SCAD, [8]) and minimax concavity penalty (MCP, [10]). Figure 4 visualizes these penalty functions for $\lambda = 1$. We see that all penalty functions are folded concave, but the soft-thresholding ($L_1$-)penalty is also convex. The parameter $\gamma$ in SCAD and MCP controls the degree of concavity. From Fig. 4c and d, we see that a smaller value of $\gamma$ results in more concave penalties. When $\gamma$ becomes larger, SCAD and MCP converge to the soft-thresholding penalty. MCP is a generalization of the hard-thresholding penalty which corresponds to $\gamma = 1$.

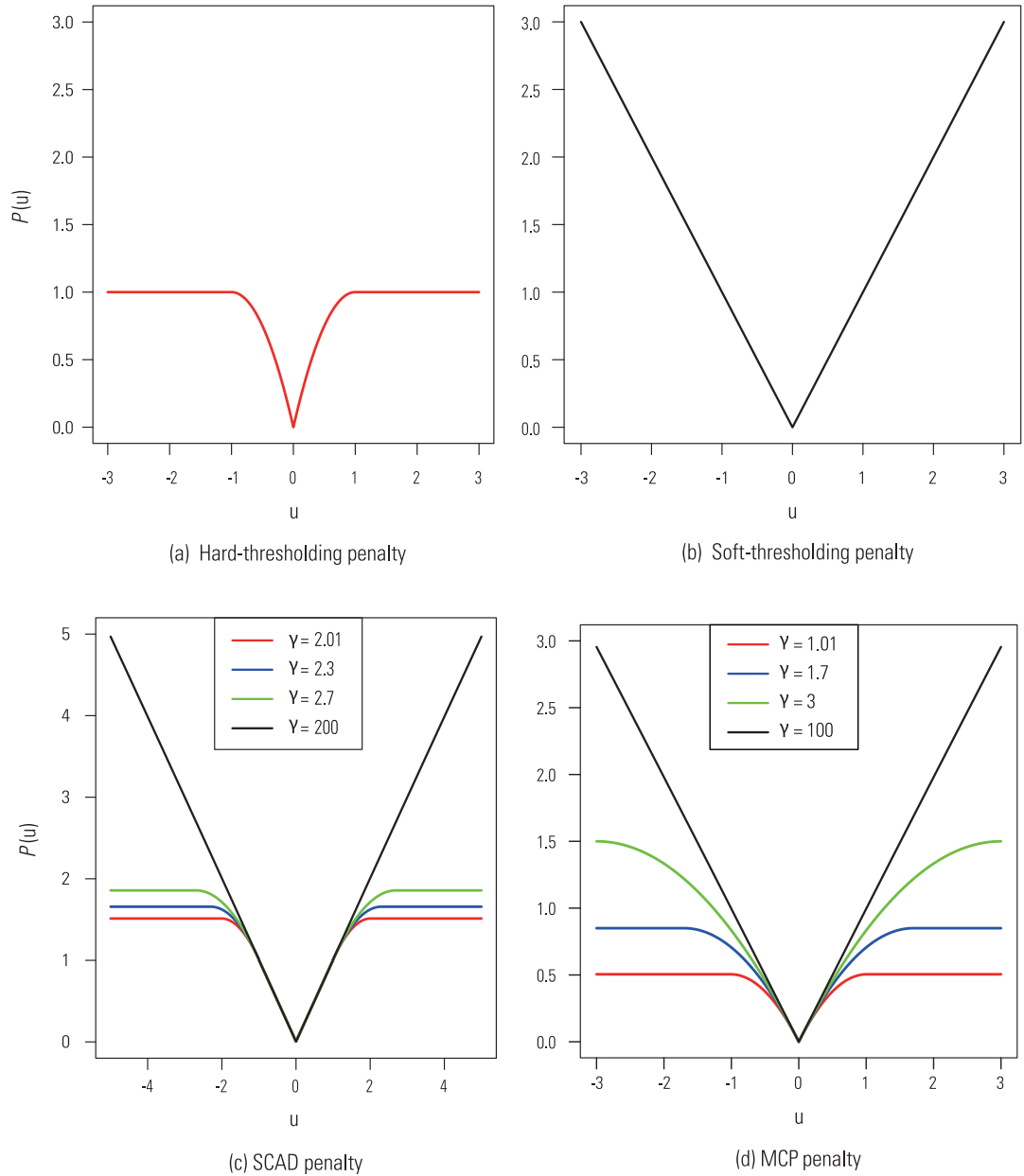

**Figure 4.** Visualization of the penalty functions. In all cases, $\lambda = 1$. For SCAD and MCP, different values of $\gamma$ are chosen as shown in graphs.

How shall we choose among these penalty functions? In applications, we recommend to use either SCAD or MCP thresholding, since they combine the advantages of both hard- and soft-thresholding operators. Many efficient algorithms have been proposed for solving the optimization problem in (9) with the above four penalties. See the 'Impact on computing infrastructure' section.

## Sparsest solution in high confidence set

The penalized quasi-likelihood estimator (9) is somewhat mysterious. A closely related method is the sparsest solution in high confidence set, introduced in the recent book chapter by [17], which has much better statistical intuition. It is a generally applicable principle that separates the data information and the sparsity assumption.

Suppose that the data information is summarized by the function $\ell_n(\boldsymbol{\beta})$ in (9). This can be a likelihood, quasi-likelihood or loss function. The underlying parameter vector $\boldsymbol{\beta}_0$ usually satisfies $\ell'(\boldsymbol{\beta}_0) = 0$, where $\ell'(\cdot)$ is the gradient vector of the expected loss function $\ell(\boldsymbol{\beta}) = \mathbb{E}\ell_n(\boldsymbol{\beta})$. Thus, a natural confidence set for $\boldsymbol{\beta}_0$ is

$$\mathcal{C}_n = \{\boldsymbol{\beta} \in \mathbb{R}^d : \|\ell'_n(\boldsymbol{\beta})\|_\infty \leq \gamma_n\}, \qquad (10)$$

where $\|\cdot\|_\infty$ is the $L_\infty$-norm of a vector and $\gamma_n$ is chosen, so that we have confidence level at least $1 - \delta_n$, namely

$$\mathbb{P}(\boldsymbol{\beta}_0 \in \mathcal{C}_n) = \mathbb{P}\{\|\ell'_n(\boldsymbol{\beta}_0)\|_\infty \leq \gamma_n\} \geq 1 - \delta_n. \quad (11)$$

The confidence set $\mathcal{C}_n$ is called high-confidence set since $\delta_n \to 0$. In theory, we can take any norm in constructing the high-confidence set. We opt for the $L_\infty$ norm, as it produces a convex confidence set $\mathcal{C}_n$ when $\ell_n(\cdot)$ is convex.

The high-confidence set is a summary of the information we have for the parameter vector $\boldsymbol{\beta}_0$. It is not informative in high-dimensional space. Take, for example, the linear model (3) with the quadratic loss $\ell_n(\boldsymbol{\beta}) = \|\mathbf{y} - \mathbf{X}\boldsymbol{\beta}\|_2^2$. The high-confidence set is then

$$\mathcal{C}_n = \{\boldsymbol{\beta} \in \mathbb{R}^d : \|\mathbf{X}^T(\mathbf{y} - \mathbf{X}\boldsymbol{\beta})\|_\infty \leq \gamma_n\},$$

where we take $\gamma_n \geq \|\mathbf{X}^T\boldsymbol{\varepsilon}\|_\infty$, so that $\delta_n = 0$. If in addition $\boldsymbol{\beta}_0$ is assumed to be sparse, then a natural solution is the intersection of these two pieces of information, namely, finding the sparsest solution in the high-confidence set:

$$\min_{\boldsymbol{\beta} \in \mathcal{C}_n} \|\boldsymbol{\beta}\|_1 = \min_{\|\ell'_n(\boldsymbol{\beta})\|_\infty \leq \gamma_n} \|\boldsymbol{\beta}\|_1. \quad (12)$$

This is a convex optimization problem when $\ell(\cdot)$ is convex. For the linear model with the quadratic loss, it reduces to the Dantzig selector [9].

There are many flexibilities in defining the sparsest solution in high-confidence set. First of all, we have a choice of the loss function $\ell_n(\cdot)$. We can regard $\ell'_n(\boldsymbol{\beta}) = 0$ as the estimation equations [83] and define directly the high-confidence set (10) from the estimation equations. Secondly, we have many ways to measure the sparsity. For example, we can use a weighted $L_1$-norm to measure the sparsity of $\boldsymbol{\beta}$ in (12). By proper choices of estimating equations in (10) and measure of sparsity in (12), the authors of [17] showed that many useful procedures can be regarded as the sparsest solution in the high-confidence set. For example, CLIME [84] for estimating sparse precision matrix in both the Gaussian graphic model and the linear programming discriminant rule [85] for sparse high-dimensional classification is the sparsest solution in the high-confidence set. The authors of [17] also provided a general convergence theory for such a procedure under a condition similar to the restricted eigenvalue condition in [86]. Finally, the idea is applicable to the problems with measurement errors or even endogeneity. In this case, the high-confidence set will be defined accordingly to accommodate the measurement errors or endogeneity. See, for example, [87].

## Independence screening

An effective variable screening technique based on marginal screening has been proposed by the authors of [11]. They aim at handling ultra-high-dimensional data for which the aforementioned penalized quasi-likelihood estimators become computationally infeasible. For such cases, the authors of [11] proposed to first use marginal regression to screen variables, reducing the original large-scale problem to a moderate-scale statistical problem, so that more sophisticated methods for variable selection can be applied. The proposed method, named sure independence screening, is computationally very attractive. It has been shown to possess sure screening property and to have some theoretical advantages over Lasso [13,88].

There are two main ideas of sure independent screening: (i) it uses the marginal contribution of a covariate to probe its importance in the joint model; and (ii) instead of selecting the most important variables, it aims at removing variables that are not important. For example, assuming each covariate has been standardized, we denote $\widehat{\beta}_j^M$ the estimated regression coefficient in a univariate regression model. The set of covariates that survive the marginal screening is defined as

$$\widehat{S} = \{j : |\widehat{\beta}_j^M| \geq \delta\} \quad (13)$$

for a given threshold $\delta$. One can also measure the importance of a covariate $X_j$ by using its deviance reduction. For the least-squares problem, both methods reduce to ranking importance of predictors by using the magnitudes of their marginal correlations with the response $Y$. The authors of [11] and [88] gave conditions under which sure screening property can be established and false selection rates are controlled.

Since the computational complexity of sure screening scales linearly with the problem size, the idea of sure screening is very effective in the dramatic reduction of the computational burden of Big Data analysis. It has been extended in various directions. For example, generalized correlation screening was used in [12], nonparametric screening was proposed by [89] and principled sure independence screening was introduced in [90]. In addition, the authors of [91] utilized the distance correlation to conduct screening, [92] employed rank correlation and [28] proposed an iteratively screening and selection method.

Independent screening has never examined the multivariate effect of variables on the response variable nor has it used the covariance matrix of variables. An extension of this is to use multivariate screening, which examines the contributions of small groups of variables together. This allows us to examine the synergy of small groups of variables to the response variable. However, the bivariate screening already involves $O(d^2)$ submodels, which can be prohibitive in computation. Covariance assist screening and estimation in [93] can be adapted here to prevent examining all bivariate or multivariate submodels. Another possible extension is to develop conditional screening techniques, which rank variables according to their conditional contributions given a set of variables.

## Dealing with incidental endogeneity

Big Data are prone to incidental endogeneity that makes the most popular regularization methods invalid. It is accordingly important to develop methods that can handle endogeneity in high dimensions. More specifically, let us consider the high-dimensional linear regression model (7). The authors of [16] showed that for any penalized estimators to be variable selection consistent, a necessary condition is

$$\mathbb{E}(\varepsilon X_j) = 0 \quad \text{for} \quad j = 1, \ldots, d. \quad (14)$$

As discussed in the 'Salient features of Big Data' section, the condition in (14) is too restrictive for real-world applications. Letting $S = \{j: \beta_j \neq 0\}$ be the set of important variables, with non-vanishing components in $\beta$, a more realistic model assumption should be

$$\mathbb{E}(\varepsilon | \{X_j\}_{j \in S}) = \mathbb{E}\Big(Y - \sum_{j \in S} \beta_j X_j | \{X_j\}_{j \in S}\Big)$$
$$= 0. \quad (15)$$

In the paper by the authors of [16], they considered an even weaker version of Equation (15), called the 'over identification' condition, such as

$$\mathbb{E}\varepsilon X_j = 0 \quad \text{and} \quad \mathbb{E}\varepsilon X_j^2 = 0 \quad \text{for } j \in S. \quad (16)$$

Under condition (16), the authors of [16] showed that the classical penalized least-squares methods, such as Lasso, SCAD and MCP, are no longer consistent. Instead, they introduced the focused generalized methods of moments (FGMMs) by utilizing the over identification conditions and proved that the FGMM consistently selects the set of variables $S$. We do not go into the technical details here but illustrate this by an example.

We continue to explore the gene expression data in the 'Incidental endogeneity' section. We again treat gene DDR1 as response and other genes as predictors, and apply the FGMM instead of Lasso. By cross validation, the FGMM selects 18 genes. The left panel of Fig. 5 shows the distribution of the sample correlations between the genes $X_j (j = 1, \ldots, 12\,718)$ and the residuals $\widehat{\varepsilon}$ after the FGMM fit. Here we find that many correlations are nonzero, but it does not matter, because we require only (16). To verify (16), the right panel of Fig. 5 shows the distribution of the sample correlations between the 18 selected genes (and their squares) and the residuals. The sample correlations between the selected genes and residuals are zero, and the sample correlations between the squared covariates and residuals are small. Therefore, the modeling assumption is consistent to our model diagnostics.

## IMPACT ON COMPUTING INFRASTRUCTURE

The massive sample size of Big Data fundamentally challenges the traditional computing infrastructure. In many applications, we need to analyze internet-scale data containing billions or even trillions of data points, which even makes a linear pass of the whole dataset unaffordable. In addition, such data could be highly dynamic and infeasible to be stored in a centralized database. The fundamental approach to store and process such data is to divide and conquer. The idea is to partition a large problem into more tractable and independent subproblems. Each subproblem is tackled in parallel by different processing units. Intermediate results from each individual worker are then combined to yield the final output. In small scale, such divide-and-conquer paradigm can be implemented either by multi-core computing or grid computing. However, in very large scale, it poses fundamental challenges to computing infrastructure. For example, when millions of computers are connected to scale out to large computing tasks, it is quite likely some computers may die during the computing. In addition, given a large computing task, we want to distribute it evenly to many computers and make the workload balanced. Designing very large scale, high adaptive and fault-tolerant computing systems is extremely challenging and motivates the outcome of new and reliable computing infrastructure that supports massively parallel data storage and processing. In this section, we take Hadoop as an example to introduce basic software and programming infrastructure for Big Data processing.

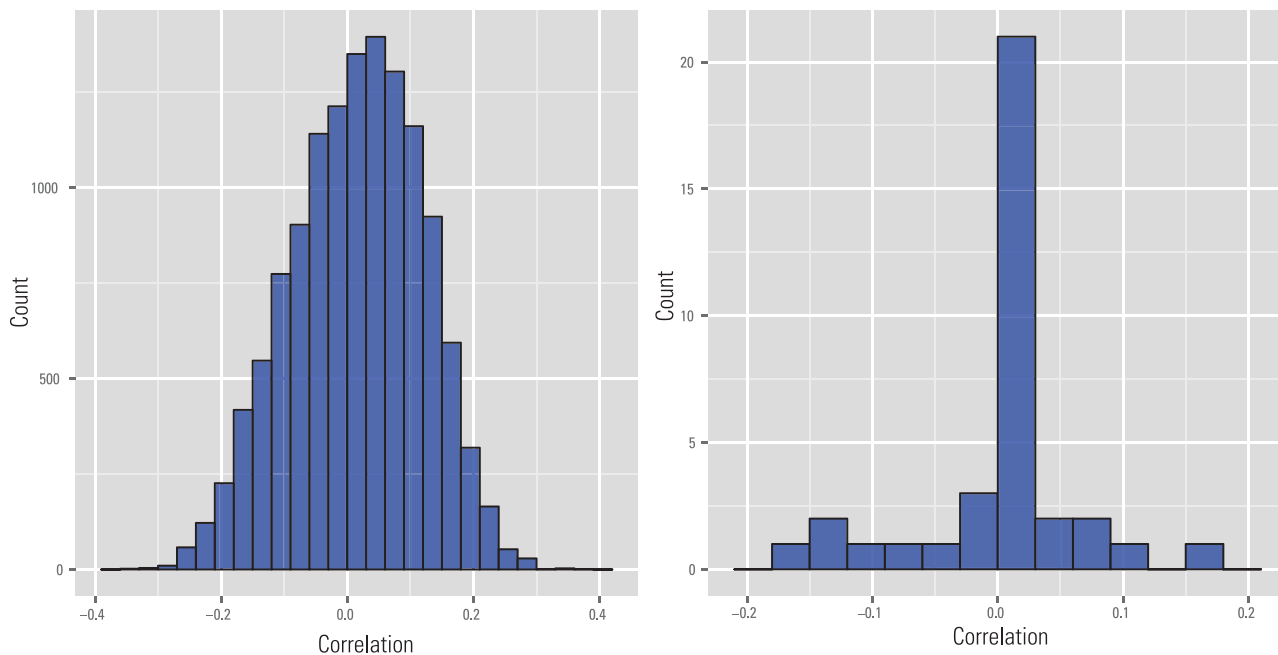

**Figure 5.** Diagnostics of the modeling assumptions of the FGMM on a microarry gene expression data. Left panel: Distribution of the sample correlations $\widehat{\text{Corr}}(X_j, \widehat{\varepsilon})$ ($j = 1, \ldots, 12{,}718$). Right panel: Distribution of the sample correlations $\widehat{\text{Corr}}(X_j, \widehat{\varepsilon})$ and $\widehat{\text{Corr}}(X_j^2, \widehat{\varepsilon})$ for only 18 selected genes. Here $\widehat{\varepsilon}$ represents the residual noise after the FGMM fit.

Hadoop is a Java-based software framework for distributed data management and processing. It contains a set of open source libraries for distributed computing using the MapReduce programming model and its own distributed file system called HDFS. Hadoop automatically facilitates scalability and takes cares of detecting and handling failures. Core Hadoop has two key components:

**Core Hadoop = Hadoop distributed file system (HDFS) + MapReduce**

- HDFS is a self-healing, high-bandwidth, clustered storage file system, and
- MapReduce is a distributed programming model developed by Google.

We dart with explaining HDFS and MapReduce in the following two subsections. Besides these two key components, a typical Hadoop release contains many other components. For example, as is shown in Fig. 6, Cloudera's open-source Hadoop distribution also includes HBase, Hive, Pig, Oozie, Flume and Sqoop. More details about these extra components are provided in the online Cloudera technical documents. After introducing the Hadoop, we also briefly explain the concepts of cloud computing in the 'Cloud computing' section.

### Hadoop distributed file system

HDFS is a distributed file system designed to host and provide high-throughput access to large datasets which are redundantly stored across multiple machines. In particular, it ensures Big Data's durability to failure and high availability to parallel applications.

As a motivating application, suppose we have a large data file containing billions of records, and we want to query this file frequently. If many queries are submitted simultaneously (e.g. the Google search engine), the usual file system is not suitable due to the I/O limit. HDFS solves this problem by dividing a large file into small blocks and store them in different machines. Each machine is called a DataNode. Unlike most block-structured file systems which use a block size on the order of 4 or 8 KB, the default block size in HDFS is 64MB, which allows HDFS to reduce the amount of metadata storage required per file. Furthermore, HDFS allows for fast streaming reads of data by keeping large amounts of data sequentially laid out on the hard disk. The main tradeoff of this decision is that HDFS expects the data to be read sequentially (instead of being read in a random access fashion).

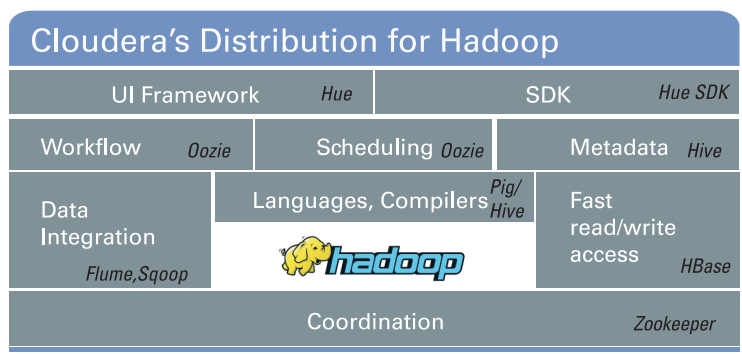

**Figure 6.** An illustration of Cloudera's open-source Hadoop distribution (source: cloudera website).

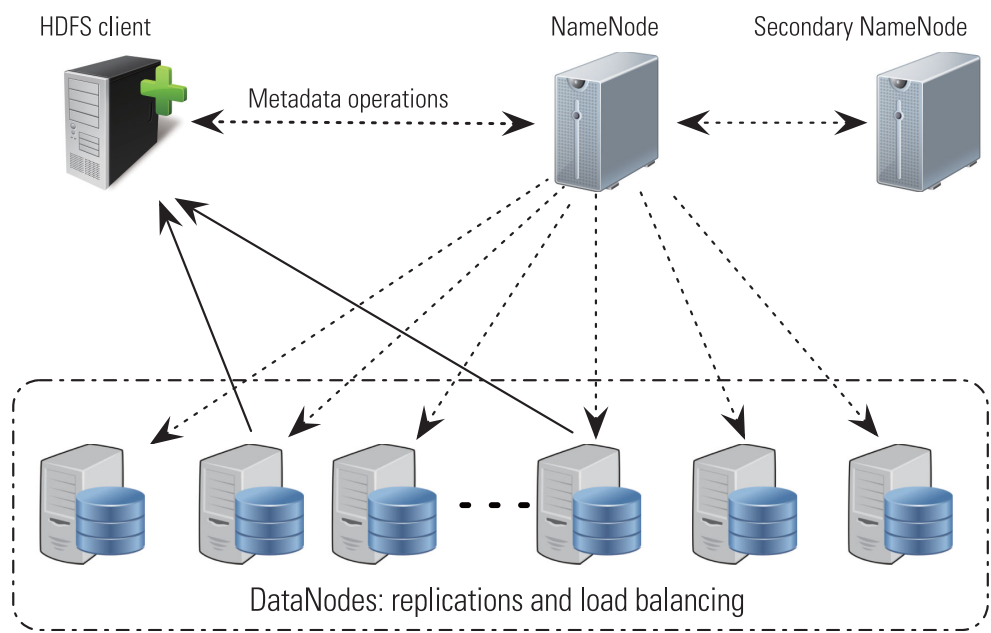

**Figure 7.** An illustration of the HDFS architecture.

The data in HDFS can be accessed via a 'write once and read many' approach. The metadata structures (e.g. the file names and directories) are allowed to be simultaneously modified by many clients. It is important that this meta information is always synchronized and stored reliably. All the metadata are maintained by a single machine, called the NameNode. Because of the relatively low amount of metadata per file (it only tracks file names, permissions and the locations of each block of each file), all such information can be stored in the main memory of the NameNode machine, allowing fast access to the metadata. An illustration of the whole HDFS architecture is provided in Fig. 7.

To access or manipulate a data file, a client contacts the NameNode and retrieves a list of locations for the blocks that comprise the file. These locations identify the DataNodes which hold each block. Clients then read file data directly from the DataNode servers, possibly in parallel. The NameNode is not directly involved in this bulk data transfer, keeping its working load to a minimum. HDFS has a built-in redundancy and replication feature which secures that any failure of individual machines can be recovered without any loss of data (e.g. each DataNode has three copies by default). The HDFS automatically balances its load whenever a new DataNode is added to the cluster. We also need to safely store the NameNode information by creating multiple redundant systems, which allows the important metadata of the file system be recovered even if the NameNode itself crashes.

## MapReduce

MapReduce is a programming model for processing large datasets in a parallel fashion. We use an example to explain how MapReduce works. Suppose we are given a symbol sequence (e.g. 'ATGCCAATC-GATGGGACTCC'), and the task is to write a program that counts the number of each symbol. The simplest idea is to read a symbol, add it into a hash table with key as the symbol and set value to its number of occurrences. If the symbol is not in the hash table yet, then add the symbol as a new key to the hash and set the corresponding value to 1. If the symbol is already in the hash table, then increase the value by 1. This program runs in a serial fashion and the time complexity scales linearly with the length of the symbol sequence. Everything looks simple so far. However, imagine if instead of a simple sequence, we need to count the number of symbols in the whole genomes of many biological subjects. Serial processing of such a huge amount of information is time consuming. So, the question is how can we use parallel processing units speed up the computation.

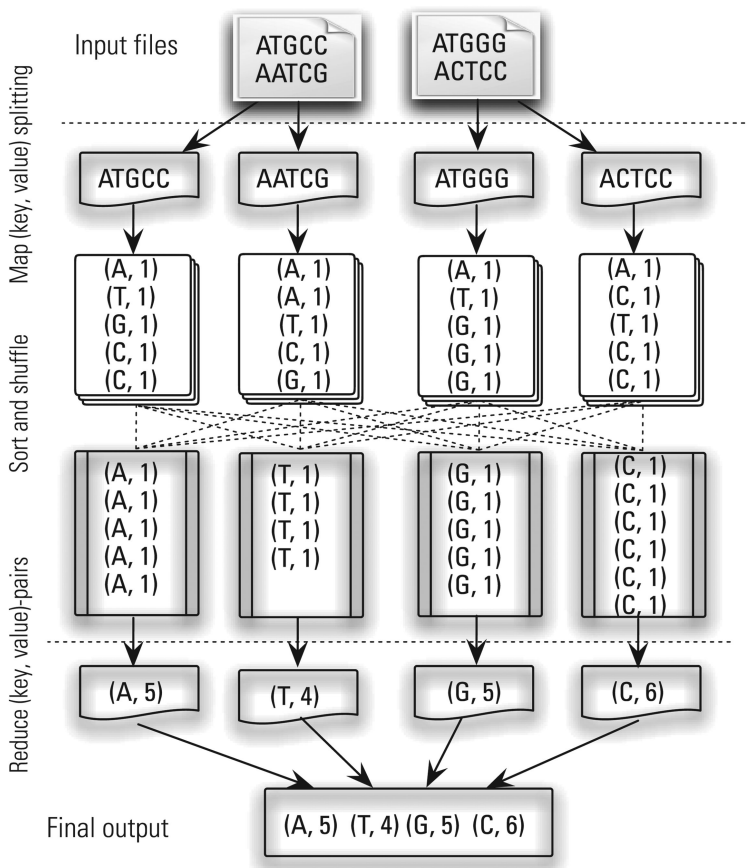

**Figure 8.** An illustration of the MapReduce paradigm for the symbol counting task. Mappers are applied to every element of the input sequences and emit intermediate (key, value)-pairs. Reducers are applied to all values associated with the same key. Between the map and reduce stages are some intermediate steps involving distributed sorting and grouping.

The idea of MapReduce is illustrated in Fig. 8. We initially split the original sequence into several files (e.g. two files in this case). We further split each file into several subsequences (e.g. two subsequences in this case) and 'map' the number of each symbol in each subsequence. The outputs of the mapper are (key, value)-pairs. We then gather together all output pairs of the mappers with the same key. Finally, we use a 'reduce' function to combine the values for each key. This gives the desired output:

$$\#A = 5, \#T = 4, \#G = 5, \#C = 6.$$

The Hadoop MapReduce contains three stages, which are listed as follows.

*First stage: mapping.* The first stage of a MapReduce program is called mapping. In this stage, a list of data elements is provided to a 'mapper' function to be transformed into (key, value)-pairs. For example, in the above symbol-counting problem, the mapper function simply transforms each symbol into the pair (symbol, 1). The mapper function does not modify the input data, but simply returns a new output list.

*Intermediate stages: shuffling and sorting.* After the mapping stage, the program exchanges the intermediate outputs from the mapping stage to different 'reducers'. This process is called shuffling. A different subset of the intermediate key space is assigned to each reduce node. These subsets (known as 'partitions') are the inputs to the next reducing step. Each map task may send (key, value)-pairs to any partition. All pairs with the same key are always grouped together on the same reducer regardless of which mappers they are coming from. Each reducer may process several sets of pairs with different keys. In this case, different keys on a single node are automatically sorted before they are fed into the next reducing step.

*Final stage: reducing.* In the final reducing stage, an instance of a user-provided code is called for each key in the partition assigned to a reducer. The inputs are a key and an iterator over all the values associated with the key. These values returned by the iterator could be in an undefined order. In particular, we have one output file per executed reduce task.

The Hadoop MapReduce builds on the HDFS and inherits all the fault-tolerance properties of HDFS. In general, Hadoop is deployed on very large scale clusters. One example is shown in Fig. 9.

## Cloud computing

Cloud computing revolutionizes modern computing paradigm. It allows everything—from hardware resources, software infrastructure to datasets—to be delivered to data analysts as a service wherever and whenever needed. Figure 10 illustrates different building components of cloud computing. The most striking feature of cloud computing is its elasticity and ability to scale up and down, which makes it suitable for storing and processing Big Data.

## IMPACT ON COMPUTATIONAL METHODS

Big Data are massive and very high dimensional, which pose significant challenges on computing and paradigm shifts on large-scale optimization [29,94]. On the one hand, the direct application of penalized quasi-likelihood estimators on high-dimensional data requires us to solve very large scale optimization problems. Optimization with a large amount of variables is not only expensive but also scribed in the 'Impact on statistical thinking' section. We will introduce scalable first-order algorithms for solving these estimators in the 'First-order methods for nonsmooth optimization' section. We also note that the volumes of modern datasets are exploding and it is often computationally infeasible to directly make inference based on the raw data. Accordingly, to effectively handle Big Data in both statistical and computational perspectives, dimension reduction as an important data pre-processing step is advocated and exploited in many applications [95]. We will explain some effective dimension reduction methods in the 'Dimension reduction and random projection' section.

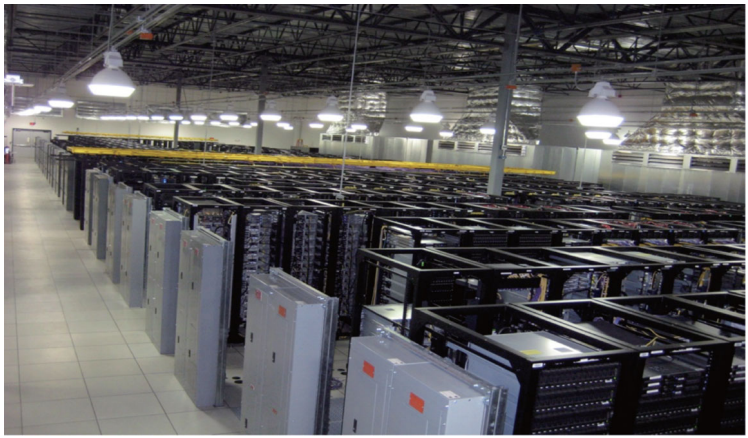

**Figure 9.** A typical Hadoop cluster (source: wikipedia).

suffers from slow numerical rates of convergence and instability. Such a large-scale optimization is generally regarded as a mean, not the goal of Big Data analysis. Scalable implementations of large-scale nonsmooth optimization procedures are crucially needed. On the other hand, the massive sample size of Big Data, which can be in the order of millions or even billions as in genomics, neuroinformatics, marketing, and online social medias, also gives rise to intensive computation on data management and queries. Parallel computing, randomized algorithms, approximate algorithms and simplified implementations should be sought. Therefore, the scalability of statistical methods to both high dimensionality and large sample size should be seriously considered in the development of statistical procedures.

In this section, we explain some new progress on developing computational methods that are scalable to Big Data. To balance the statistical accuracy and computational efficiency, several penalized estimators such as Lasso, SCAD, and MCP have been de-

### First-order methods for nonsmooth optimization

In this subsection, we introduce several first-order optimization algorithms for solving the penalized quasi-likelihood estimators in (9). For most loss functions $\ell_n(\cdot)$, this optimization problem has no closed-form solution. Iterative procedures are needed to solve it.

When the penalty function $P_{\lambda,\gamma}(\cdot)$ is convex (e.g. the $L_1$-penalty), so is the objective function in (9) when $\ell_n(\cdot)$ is convex. Accordingly, sophisticated convex optimization algorithms can be applied. The most widely used convex optimization algorithm is gradient descent [96], which finds a solution sequence converging to the optimum $\widehat{\boldsymbol{\beta}}$ by calculating the gradient of the objective function at each point. However, calculating the gradient can be very time consuming when the dimensionality is high. Instead, the authors of [97] proposed to calculate the penalized pseudo-likelihood estimator using the pathwise coordinate descent algorithm, which can be viewed as a special case of the gradient descent algorithm. Instead of optimizing along the direction of the full gradient, it only calculates the gradient direction along one coordinate at each time. A beautiful feature of this is that even though the whole optimization problem does not have a closed-form solution, there exist simple closed-form solutions to all the univariate subproblems. The coordinate descent is computationally easy and has similar numerical convergence properties as gradient descent [98]. Alternative first-order algorithms to coordinate descent have also been proposed and widely used, resulting in iterative shrinkage-thresholding algorithms [23,24]. Prior to the coordinate descent algorithm, the authors of [19] proposed the least angle regression (LARS) algorithm to the $L_1$-penalized least-squares problem.

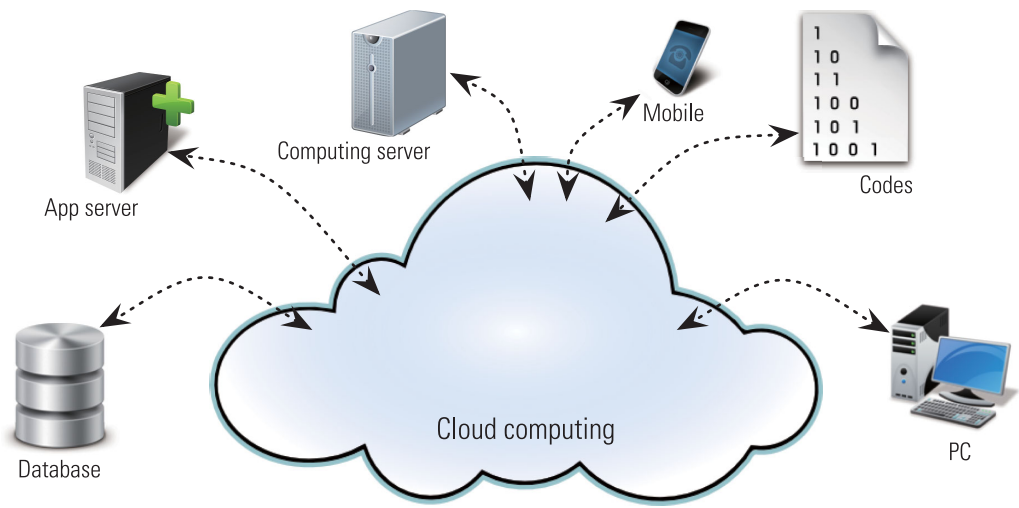

**Figure 10.** An illustration of the cloud computing paradigm.

When the penalty function $P_{\lambda,\gamma}(\cdot)$ is nonconvex (e.g. SCAD and MCP), the objective function in (9) is no longer concave. Many algorithms have been proposed to solve this optimization problem. For example, the authors of [8] proposed a local quadratic approximation (LQA) algorithm for optimizing nonconcave penalized likelihood. Their idea is to approximate the penalty term piece by piece using a quadratic function, which can be thought as a convex relaxation (majorization) to the nonconcave object function. With the quadratic approximation, a closed-form solution can be obtained. This idea is further improved by using a linear instead of a quadratic function to approximate the penalty term and leads to the local linear approximation (LLA) algorithm [27]. More specifically, given current estimate $\widehat{\boldsymbol{\beta}}^{(k)} = (\beta_1^{(k)}, \ldots, \beta_d^{(k)})^T$ at the $k$th iteration for problem (9), by Taylor's expansion,

$$P_{\lambda,\gamma}(\beta_j) \approx P_{\lambda,\gamma}\left(\beta_j^{(k)}\right)$$
$$+ P'_{\lambda,\gamma}\left(\beta_j^{(k)}\right)\left(|\beta_j| - |\beta_j^{(k)}|\right). \quad (17)$$

Thus, at the $(k+1)$th iteration, we solve

$$\min_{\beta_j} \left\{ \ell_n(\boldsymbol{\beta}) + \sum_{j=1}^d w_{k,j}|\beta_j| \right\}, \quad (18)$$

where $w_{k,j} = P'_{\lambda,\gamma}(\beta_j^{(k)})$. Note that problem (18) is convex, so that a convex solver can be used. The authors of [58] suggested using initial values $\boldsymbol{\beta}^{(0)} = \mathbf{0}$, which corresponds to the unweighted $L_1$ penalty. This algorithm shares a very similar idea as in [99], both of which can be regarded as implementations of the minimization of the folded-concave penalized quasi-likelihood [8] problem (9). If one further approximates the goodness-of-fit measure $\ell_n(\boldsymbol{\beta})$ in (18) by a quadratic function via the Taylor expansion, then the LARS algorithm [19] and pathwise coordinate descent algorithm [97] can be used.

For the more general settings where the loss function $\ell_n(\cdot)$ may not be concave, the authors of [100] proposed an approximate regularization path following algorithm for solving the optimization problem in (9). By integrating statistical analysis with computational algorithms, they provided explicit statistical and computational rates of convergence of any local solution obtained by the algorithm. Computationally, the approximate regularization path following algorithm attains a global geometric rate of convergence for calculating the full regularization path, which is fastest possible among all first-order algorithms in terms of iteration complexity. Statistically, they show that any local solution obtained by the algorithm attains the oracle properties with the optimal rates of convergence. The idea on studying statistical properties based on computational algorithms, which combine both computational and statistical analysis, represents an interesting future direction for Big Data. We also refer to [101] and [102] for research studies in this direction.

## Dimension reduction and random projection

We introduce several dimension (data) reduction procedures in this section. Why do we need dimension reduction? Let us consider a dataset represented as an $n \times d$ real-value matrix $\mathbf{D}$, which encodes information about $n$ observations of $d$ variables. In the Big Data era, it is in general computationally intractable to directly make inference on the raw data matrix. Therefore, an important data-preprocessing procedure is to conduct dimension reduction which finds a compressed representation of $\mathbf{D}$ that is of lower dimensions but preserves as much information in $\mathbf{D}$ as possible.

Principal component analysis (PCA) is the most well-known dimension reduction method. It aims at projecting the data onto a low-dimensional orthogonal subspace that captures as much of the data variation as possible. Empirically, it calculates the leading eigenvectors of the sample covariance matrix to form a subspace $\widehat{\mathbf{U}}_k \in \mathbb{R}^{d \times k}$. We then project the $n \times d$ data matrix $\mathbf{D}$ to this linear subspace to obtain an $n \times k$ data matrix $\mathbf{D}\widehat{\mathbf{U}}_k$. This procedure is optimal among all the linear projection methods in minimizing the squared error introduced by the projection. However, conducting the eigenspace decomposition on the sample covariance matrix is computational challenging when both $n$ and $d$ are large. The computational complexity of PCA is $O(d^2n + d^3)$ [103], which is infeasible for very large datasets.

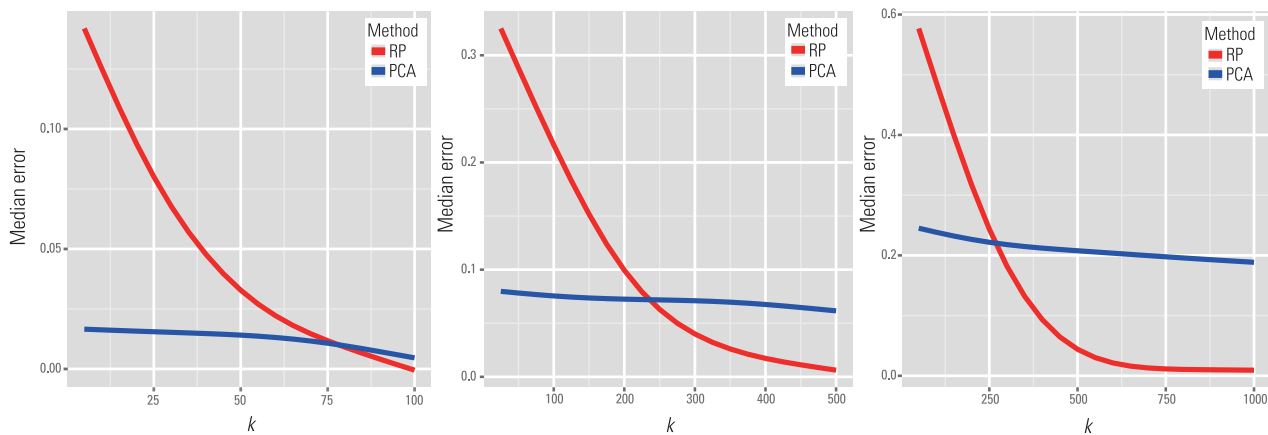

**Figure 11.** Plots of the median errors in preserving the distances between pairs of data points versus the reduced dimension *k* in large-scale microarray data. Here 'RP' stands for the random projection and 'PCA' stands for the principal component analysis.

To handle the computational challenge raised by massive and high-dimensional datasets, we need to develop methods that preserve the data structure as much as possible and is computational efficient for handling high dimensionality. Random projection (RP) [104] is an efficient dimension reduction technique for this purpose, and is closely related to the celebrated idea of compress sensing [105–109]. More specifically, RP aims at finding a *k*-dimensional subspace of **D**, such that the distances between all pairs of data points are approximately preserved. It achieves this goal by projecting the original data **D** onto a *k*-dimensional subspace using an RP matrix with unit column norms. More specifically, let $\mathbf{R} \in \mathbb{R}^{d \times k}$ be a random matrix with all the column Euclidean norms equal to 1. We reduce the dimensionality of **D** from *d* to *k* by calculating matrix multiplication

$$\widehat{\mathbf{D}}^R = \mathbf{DR}.$$

This procedure is very simple and the computational complexity of the RP procedure is of order $O(ndk)$, which scales only linearly with the problem size.

Theoretical justifications of RP are based on two results. The authors of [104] showed that if points in a vector space are projected onto a randomly selected subspace of suitable dimensions, then the distances between the points are approximately preserved. This justifies the RP when **R** is indeed a projection matrix. However, enforcing **R** to be orthogonal requires the Gram–Schmidt algorithm, which is computationally expensive. In practice, the authors of [110] showed that in high dimensions we do not need to enforce the matrix to be orthogonal. In fact, any finite number of high-dimensional random vectors are almost orthogonal to each other. This result guarantees that $\mathbf{R}^T \mathbf{R}$ can be sufficiently close to the identity matrix. The authors of [111] further simplified the RP procedure by removing the unit column length constraint.

To illustrate the usefulness of RP, we use the gene expression data in the 'Incidental endogeneity' section to compare the performance of PCA and RP in preserving the relative distances between pairwise data points. We extract the top 100, 500 and 2500 genes with the highest marginal standard deviations, and then apply PCA and RP to reduce the dimensionality of the raw data to a small number *k*. Figure 11 shows the median errors in the distance between members across all pairs of data vectors. We see that, when dimensionality increases, RPs have more and more advantages over PCA in preserving the distances between sample pairs.

One thing to note is that RP is not the 'optimal' procedure for traditional small-scale problems. Accordingly, the popularity of this dimension reduction procedure indicates a new understanding of Big Data. To balance the statistical accuracy and computational complexity, the suboptimal procedures in small- or medium-scale problems can be 'optimal' in large scale. Moreover, the theory of RP depends on the high dimensionality feature of Big Data. This can be viewed as a blessing of dimensionality.

Besides PCA and RP, there are many other dimension-reduction methods, including latent semantic indexing (LSI) [112], discrete cosine transform [113] and CUR decomposition [114]. These methods have been widely used in analyzing large text and image datasets.

## CONCLUSIONS AND FUTURE PERSPECTIVES

This paper discusses statistical and computational aspects of Big Data analysis. We selectively overview several unique features brought by Big Data and discuss some solutions. Besides the challenge of massive sample size and high dimensionality, there are several other important features of Big Data worth equal attention. These include

(1) Complex data challenge: due to the fact that Big Data are in general aggregated from multiple sources, they sometime exhibit heavy tail behaviors with nontrivial tail dependence.
(2) Noisy data challenge: Big Data usually contain various types of measurement errors, outliers and missing values.
(3) Dependent data challenge: in various types of modern data, such as financial time series, fMRI and time course microarray data, the samples are dependent with relatively weak signals.

To handle these challenges, it is urgent to develop statistical methods that are robust to data complexity (see, for example, [115–117]), noises [62–119] and data dependence [51,120–122].

## ACKNOWLEDGEMENTS

The authors gratefully acknowledge Dr Emre Barut for his kind assistance on producing Fig. 5. The authors thank the associate editor and referees for helpful comments.

## FUNDING

This work was supported by the National Science Foundation [DMS-1206464 to JQF, III-1116730 and III-1332109 to HL] and the National Institutes of Health [R01-GM100474 and R01-GM072611 to JQF].

## REFERENCES

1. Stein, L. The case for cloud computing in genome informatics. *Genome Biol* 2010; **11**: 207.
2. Donoho, D. High-dimensional data analysis: the curses and blessings of dimensionality. In: *The American Mathematical Society Conference*, Los Angeles, CA, United States, 7–12 August 2000.
3. Bickel, P. Discussion on the paper 'Sure independence screening for ultrahigh dimensional feature space' by Fan and Lv. *J Roy Stat Soc B* 2008; **70**: 883–4.
4. Fan, J and Fan, Y. High dimensional classification using features annealed independence rules. *Ann Stat* 2008; **36**: 2605–37.
5. Pittelkow, PH and Ghosh, M. Theoretical measures of relative performance of classifiers for high dimensional data with small sample sizes. *J Roy Stat Soc B* 2008; **70**: 159–73.
6. Tibshirani, R. Regression shrinkage and selection via the lasso. *J Roy Stat Soc B* 1996; **58**: 267–88.
7. Chen, S, Donoho, D and Saunders, M. Atomic decomposition by basis pursuit. *SIAM J Sci Comput* 1998; **20**: 33–61.
8. Fan, J and Li, R. Variable selection via nonconcave penalized likelihood and its oracle properties. *J Am Stat Assoc* 2001; **96**: 1348–60.
9. Candes, E and Tao, T. The Dantzig selector: statistical estimation when *p* is much larger than *n*. *Ann Stat* 2007; **35**: 2313–51.
10. Zhang, C-H. Nearly unbiased variable selection under minimax concave penalty. *Ann Stat* 2010; **38**: 894–942.
11. Fan, J and Lv, J. Sure independence screening for ultrahigh dimensional feature space (with discussion). *J Roy Stat Soc B* 2008; **70**: 849–911.
12. Hall, P and Miller, H. Using generalized correlation to effect variable selection in very high dimensional problems. *J Comput Graph Stat* 2009; **18**: 533–50.
13. Genovese, C, Jin, J and Wasserman, L *et al.* A comparison of the lasso and marginal regression. *J Mach Learn Res* 2012; **13**: 2107–43.
14. Fan, J, Guo, S and Hao, N. Variance estimation using refitted cross-validation in ultrahigh dimensional regression. *J Roy Stat Soc B* 2012; **74**: 37–65.
15. Liao, Y and Jiang, W. Posterior consistency of nonparametric conditional moment restricted models. *Ann Stat* 2011; **39**: 3003–31.
16. Fan, J and Liao, Y. Endogeneity in ultrahigh dimension. *Technical report*. Princeton University, 2012.
17. Fan, J. Features of big data and sparsest solution in high confidence set. *Technical report*. Princeton University, 2013.
18. Donoho, D and Elad, M. Optimally sparse representation in general (nonorthogonal) dictionaries via $L_1$ minimization. *Proc Natl Acad Sci USA* 2003; **100**: 2197–202.
19. Efron, B, Hastie, T and Johnstone, I *et al.* Least angle regression. *Ann Stat* 2004; **32**: 407–99.
20. Friedman, J and Popescu, B. Gradient directed regularization for linear regression and classification. *Technical report*. Stanford University, 2003.
21. Fu, WJ. Penalized regressions: the bridge versus the lasso. *J Comput Graph Stat* 1998; **7**: 397–416.
22. Wu, T and Lange, K. Coordinate descent algorithms for lasso penalized regression. *Ann Appl Stat* 2008; **2**: 224–44.
23. Daubechies, I, Defrise, M and De Mol, C. An iterative thresholding algorithm for linear inverse problems with a sparsity constraint. *Commun Pur Appl Math* 2004; **57**: 1413–57.
24. Beck, A and Teboulle, M. A fast iterative shrinkage-thresholding algorithm for linear inverse problems. *SIAM J Imaging Sciences* 2009; **2**: 183–202.
25. Lange, K, Hunter, D and Yang, I. Optimization transfer using surrogate objective functions. *J Comput Graph Stat* 2000; **9**: 1–20.
26. Hunter, D and Li, R. Variable selection using MM algorithms. *Ann Stat* 2005; **33**: 1617–42.
27. Zou, H and Li, R. One-step sparse estimates in nonconcave penalized likelihood models. *Ann Stat* 2008; **36**: 1509–33.
28. Fan, J, Samworth, R and Wu, Y. Ultrahigh dimensional feature selection: beyond the linear model. *J Mach Learn Res* 2009; **10**: 2013–38.

29. Boyd, S, Parikh, N and Chu, E et al. Distributed optimization and statistical learning via the alternating direction method of multipliers. *Found Trends Mach Learn* 2011; **3**: 1–122.

30. Bradley, J, Kyrola, A and Bickson, D et al. Parallel coordinate descent for $L_1$-regularized loss minimization. arXiv:1105.5379, 2011.

31. Low, Y, Bickson, and Dand Gonzalez, J et al. Distributed graphlab: a framework for machine learning and data mining in the cloud. *Proc Int Conf VLDB Endowment* 2012; **5**: 716–27.

32. Worthey, E, Mayer, A and Syverson, G et al. Making a definitive diagnosis: successful clinical application of whole exome sequencing in a child with intractable inflammatory bowel disease. *Genet Med* 2010; **13**: 255–62.

33. Chen, R, Mias, G and Li-Pook-Than, J et al. Personal omics profiling reveals dynamic molecular and medical phenotypes. *Cell* 2012; **148**: 1293–307.

34. Cohen, J, Kiss, R and Pertsemlidis, A et al. Multiple rare alleles contribute to low plasma levels of HDL cholesterol. *Science* 2004; **305**: 869–72.

35. Han, F and Pan, W. A data-adaptive sum test for disease association with multiple common or rare variants. *Hum Hered* 2010; **70**: 42–54.

36. Bickel, P, Brown, J and Huang, H et al. An overview of recent developments in genomics and associated statistical methods. *Philos T R Soc A* 2009; **367**: 4313–37.

37. Leek, J and Storey, J. Capturing heterogeneity in gene expression studies by surrogate variable analysis. *PLoS Genet* 2007; **3**: e161.

38. Benjamini, Y and Hochberg, Y. Controlling the false discovery rate: a practical and powerful approach to multiple testing. *J Roy Stat Soc B* 1995; **57**: 289–300.

39. Storey, J. The positive false discovery rate: a Bayesian interpretation and the q-value. *Ann Stat* 2003; **31**: 2013–35.

40. Schwartzman, A, Dougherty, R and Lee, J et al. Empirical null and false discovery rate analysis in neuroimaging. *Neuroimage* 2009; **44**: 71–82.

41. Efron, B. Correlated z-values and the accuracy of large-scale statistical estimates. *J Am Stat Assoc* 2010; **105**: 1042–55.

42. Fan, J, Han, X and Gu, W. Control of the false discovery rate under arbitrary covariance dependence. *J Am Stat Assoc* 2012; **107**: 1019–45.

43. Edgar, R, Domrachev, M and Lash, AE. Gene expression omnibus: NCBI gene expression and hybridization array data repository. *Nucleic Acids Res* 2002; **30**: 207–10.

44. Jonides, J, Nee, D and Berman, M. What has functional neuroimaging told us about the mind? So many examples little space. *Cortex* 2006; **42**: 414–7.

45. Visscher, K and Weissman, D. Would the field of cognitive neuroscience be advanced by sharing functional MRI data? *BMC Med* 2011; **9**: 34.

46. Milham, M, Mennes, M and Gutman, D et al. The International Neuroimaging Data-sharing Initiative (INDI) and the Functional Connectomes Project. *17th Annual Meeting of the Organization for Human Brain Mapping,* Quebec City, 2011.

47. Di Martino, A., Yan, CG and Li, Q et al. The autism brain imaging data exchange: Towards a large-scale evaluation of the intrinsic brain architecture in autism. *Mol Psychiatry* 2013, doi:10.1038/mp.2013.78.

48. The ADHD-200 Consortium. The ADHD-200 consortium: a model to advance the translational potential of neuroimaging in clinical neuroscience. *Front Syst Neurosci* 2012; **6**: 62.

49. Fritsch, V, Varoquaux, G and Thyreau, B et al. Detecting outliers in high-dimensional neuroimaging datasets with robust covariance estimators. *Med Image Anal* 2012; **16**: 1359–70.

50. Song, S and Bickel, P. *Large vector auto regressions.* arXiv:1106.3915, 2011.

51. Han, F and Liu, H. Transition matrix estimation in high dimensional time series. In: *The 30th International Conference on Machine Learning*, Atlanta, GA, USA, 16–21 June, 2013.

52. Cochrane, J. *Asset Pricing.* Princeton, NJ: Princeton University Press, 2001.

53. Dempster, M. *Risk Management: Value at Risk and Beyond.* Cambridge: Cambridge University Press, 2002.

54. Stock, J and Watson, M. Forecasting using principal components from a large number of predictors. *J Am Stat Assoc* 2002; **97**: 1167–79.

55. Bai, J and Ng, S. Determining the number of factors in approximate factor models. *Econometrica* 2002; **70**: 191–221.

56. Bai, J. Inferential theory for factor models of large dimensions. *Econometrica* 2003; **71**: 135–71.

57. Forni, M, Hallin, M and Lippi, M et al. The generalized dynamic factor model: one-sided estimation and forecasting. *J Am Stat Assoc* 2005; **100**: 830–40.

58. Fan, J, Fan, Y and Lv, J. High dimensional covariance matrix estimation using a factor model. *J. Econometrics* 2008; **147**: 186–97.

59. Bickel, P and Levina, E. Covariance regularization by thresholding. *Ann Stat* 2008; **36**: 2577–604.

60. Cai, T and Liu, W. Adaptive thresholding for sparse covariance matrix estimation. *J Am Stat Assoc* 2011; **106**: 672–84.

61. Agarwal, A, Negahban, S and Wainwright, M. Noisy matrix decomposition via convex relaxation: optimal rates in high dimensions. *Ann Stat* 2012; **40**: 1171–97.

62. Liu, H, Han, F and Yuan, M et al. High-dimensional semiparametric Gaussian copula graphical models. *Ann Stat* 2012; **40**: 2293–326.

63. Xue, L and Zou, H. Regularized rank-based estimation of high-dimensional nonparanormal graphical models. *Ann Stat* 2012; **40**: 2541–71.

64. Liu, H, Han, F and Zhang, C-H. Transelliptical graphical models. In: *The 25th Conference in Advances in Neural Information Processing Systems*, Lake Tahoe, NV, USA, 3–8 December, 2012.

65. Fan, J, Liao, Y and Mincheva, M. Large covariance estimation by thresholding principal orthogonal complements. *J Roy Stat Soc B* 2013; **75**: 603–80.

66. Pourahmadi, M. *Modern Methods to Covariance Estimation: With High-Dimensional Data.* New York: Wiley, 2013.

67. Aramaki, E, Maskawa, S and Morita, M. Twitter catches the flu: detecting influenza epidemics using twitter. In: *The Conference on Empirical Methods in Natural Language Processing*, Edinburgh, UK, 27–29 July, 2011.

68. Bollen, J, Mao, H and Zeng, X. Twitter mood predicts the stock market. *J Comput Sci* 2011; **2**: 1–8.

69. Asur, S and Huberman, B. Predicting the future with social media. In: *The IEEE/WIC/ACM International Conference on Web Intelligence and Intelligent Agent Technology (WI-IAT)*, Toronto, Canada, 31 August–3 September, 2010.

70. Khalili, A and Chen, J. Variable selection in finite mixture of regression models. *J Am Stat Assoc* 2007; **102**: 1025–38.

71. Städler, N, Bühlmann, P and van de Geer, S. $\ell_1$-penalization for mixture regression models. *Test* 2010; **19**: 209–56.

72. Hastie, T, Tibshirani, R and Friedman, J. *The Elements of Statistical Learning.* Berlin: Springer, 2009.

73. Bühlmann, P and van de Geer, S. *Statistics for High-Dimensional Data: Methods, Theory and Applications.* Berlin: Springer, 2011.

74. Cai, T and Jiang, T. Phase transition in limiting distributions of coherence of high-dimensional random matrices. *J Multivariate Anal* 2012; **107**: 24–39.

75. Engle, R, Hendry, D and Richard, J-F. Exogeneity. *Econometrica* 1983; **51**: 277–304.

76. Brazma, A, Parkinson, H and Sarkans, U et al. ArrayExpress—a public repository for microarray gene expression data at the EBI. *Nucleic Acids Res* 2003; **31**: 68–71.

77. Valiathan, R, Marco, M and Leitinger, B et al. Discoidin domain receptor tyrosine kinases: new players in cancer progression. *Cancer Metastasis Rev* 2012; **31**: 295–321.

78. Akaike, H. A new look at the statistical model identification. *IEEE Trans Automat Control* 1974; **19**: 716–23.
79. Barron, A, Birgé, L and Massart, P. Risk bounds for model selection via penalization. *Probab Theory Related Fields* 1999; **113**: 301–413.
80. Antoniadis, A. Wavelets in statistics: a review. *J Ital Stat Soc* 1997; **6**: 97–130.
81. Antoniadis, A and Fan, J. Regularization of wavelet approximations. *J Am Stat Assoc* 2001; **96**: 939–55.
82. Donoho, D and Johnstone, J. Ideal spatial adaptation by wavelet shrinkage. *Biometrika* 1994; **81**: 425–55.
83. Liang, K-Y and Zeger, S. Longitudinal data analysis using generalized linear models. *Biometrika* 1986; **73**: 13–22.
84. Cai, T, Liu, W and Luo, X. A constrained $L_1$ minimization approach to sparse precision matrix estimation. *J Am Stat Assoc* 2011; **106**: 594–607.
85. Cai, T and Liu, W. A direct estimation approach to sparse linear discriminant analysis. *J Am Stat Assoc* 2011; **106**: 1566–77.
86. Bickel, P, Ritov, Y and Tsybakov, A. Simultaneous analysis of lasso and Dantzig selector. *Ann Stat* 2009; **37**: 1705–32.
87. Gautier, E and Tsybakov, A. High-dimensional instrumental variables regression and confidence sets. arXiv:1105.2454, 2011.
88. Fan, J and Song, R. Sure independence screening in generalized linear models with NP-dimensionality. *Ann Stat* 2010; **38**: 3567–604.
89. Fan, J, Feng, Y and Song, R. Nonparametric independence screening in sparse ultra-high dimensional additive models. *J Am Stat Assoc* 2011; **106**: 544–57.
90. Zhao, S and Li, Y. Principled sure independence screening for Cox models with ultra-high-dimensional covariates. *J Multivariate Anal* 2012; **105**: 397–411.
91. Li, R, Zhong, W and Zhu, L. Feature screening via distance correlation learning. *J Am Stat Assoc* 2012; **107**: 1129–39.
92. Li, G, Peng, H and Zhang, J *et al*. Robust rank correlation based screening. *Ann Stat* 2012; **40**: 1846–77.
93. Ke, T, Jin, J and Fan, J. *Covariance assisted screening and estimation*. arXiv:1205.4645, 2012.
94. Boyd, S and Vandenberghe, L. *Convex Optimization*. Cambridge: Cambridge University Press, 2004.
95. Fodor, I. A survey of dimension reduction techniques. *Technical report*. US Department of Energy, 2002.
96. Avriel, M. *Nonlinear Programming: Analysis and Methods*. New York: Courier Dover, 2003.
97. Friedman, J, Hastie, T and Höfling, H *et al*. Pathwise coordinate optimization. *Ann Appl Stat* 2007; **1**: 302–32.
98. Nesterov, Y. Efficiency of coordinate descent methods on huge-scale optimization problems. *SIAM J Optim* 2012; **22**: 341–62.
99. Candes, E, Wakin, M and Boyd, S. Enhancing sparsity by reweighted $L_1$ minimization. *J Fourier Anal Appl* 2008; **14**: 877–905.
100. Wang, Z, Liu, H and Zhang, T. *Optimal computational and statistical rates of convergence for sparse nonconvex learning problems*. arXiv:1306.4960, 2013.
101. Agarwal, A, Negahban, S and Wainwright, M. Fast global convergence of gradient methods for high-dimensional statistical recovery. *Ann Stat* 2012; **40**: 2452–82.
102. Loh, P-L and Wainwright, M. Regularized M-estimators with nonconvexity: statistical and algorithmic theory for local optima. arXiv:1305.2436, 2013.
103. Golub, G and Van Loan, C. *Matrix Computations*. Baltimore, MD: The Johns Hopkins University Press, 2012.
104. Johnson, W and Lindenstrauss, J. Extensions of Lipschitz mappings into a Hilbert space. *Contemp Math* 1984; **26**: 189–206.
105. Donoho, D. Compressed sensing. *IEEE Trans Inform Theory* 2006; **52**: 1289–306.
106. Tsaig, Y and Donoho, D. Extensions of compressed sensing. *Signal Process* 2006; **86**: 549–71.
107. Lustig, M, Donoho, D and Pauly, J. Sparse MRI: the application of compressed sensing for rapid MR imaging. *Magn Reson Med* 2007; **58**: 1182–95.
108. Figueiredo, M, Nowak, R and Wright, S. Gradient projection for sparse reconstruction: application to compressed sensing and other inverse problems. *IEEE J Sel Top Signal Process* 2007; **1**: 586–97.
109. Candes, E and Wakin, M. An introduction to compressive sampling. *Signal Process Magazine* 2008; **25**: 21–30.
110. Marks, R and Zurada, J. *Computational Intelligence: Imitating Life*. Piscataway, NJ: IEEE, 1994.
111. Achlioptas, D. Database-friendly random projections. In: *The 20th ACM SIGMOD-SIGACT-SIGART Symposium on Principles of Database Systems*, Dallas, TX, USA, 16–18 May, 2001.
112. Deerwester, S, Dumais, S and Furnas, GR. Indexing by latent semantic analysis. *J Assn Inf Sci* 1990; **41**: 391–407.
113. Rao, K, Yip, P and Britanak, V. *Discrete Cosine Transform: Algorithms, Advantages, Applications*. New York: Academic, 2007.
114. Mahoney, M and Drineas, P. CUR matrix decompositions for improved data analysis. *Proc Natl Acad Sci USA* 2009; **106**: 697–702.
115. Owen, J and Rabinovitch, R. On the class of elliptical distributions and their applications to the theory of portfolio choice. *J Finance* 1983; **38**: 745–52.
116. Blanchard, G, Kawanabe, M and Sugiyama, M *et al*. In search of non-Gaussian components of a high-dimensional distribution. *J Mach Learn Res* 2006; **7**: 247–82.
117. Han, F and Liu, H. Scale-Invariant Sparse PCA on High Dimensional Meta-elliptical Data. *J Am Stat Assoc,* doi:10.1080/01621459.2013.844699.
118. Candes, E, Li, X and Ma, Y *et al*. Robust principal component analysis? *J. ACM* 2011; **58**: 11: 1–37.
119. Loh, P-L and Wainwright, M. High-dimensional regression with noisy and missing data: provable guarantees with nonconvexity. *Ann Stat* 2012; **40**: 1637–64.
120. Lam, C and Yao, Q. Factor modeling for high-dimensional time series: inference for the number of factors. *Ann Stat* 2012; **40**: 694–726.
121. Han, F and Liu, H. Principal component analysis on non-Gaussian dependent data. In: *The 30th International Conference on Machine Learning*, Atlanta, GA, USA, 16–21 June, 2013.
122. Huang, J, Sun, T and Ying, Z *et al*. Oracle inequalities for the lasso in the Cox model. *Ann Stat* 2013; **41**: 1142–65.